\documentclass[journal]{IEEEtran}
\ifCLASSINFOpdf
\else
\fi
\usepackage{amscd}

\usepackage{amssymb}
\usepackage{amsmath}

\usepackage{latexsym}

\usepackage{textcomp}

\usepackage{mathabx}

\usepackage{multirow}
\usepackage{hhline}

\usepackage{epstopdf}
\usepackage{epsfig}
\usepackage{enumitem}
\usepackage[numbers,sort&compress]{natbib}

\usepackage{ifpdf}

\usepackage{amscd}

\usepackage{amssymb}
\usepackage{amsmath}

\usepackage{latexsym}

\usepackage{textcomp}

\usepackage{mathabx}

\usepackage{multirow}
\usepackage{hhline}

\usepackage{subfig}

\usepackage{threeparttable}

\usepackage{algorithm}

\usepackage{algorithmicx}

\usepackage{algpseudocode}

\usepackage{url}

\begin{document}
%
\title{Measurement-Adaptive Sparse Image Sampling and Recovery}
%
%
%

\author{Ali Taimori, Farokh Marvasti,~\IEEEmembership{Senior Member, IEEE}

\thanks{A. Taimori and Farokh Marvasti are with the Electrical Engineering Department, Sharif University of Technology, Tehran 14588-89694, Iran (e-mails: alitaimori@yahoo.com; marvasti@sharif.edu).
}}
\markboth{IEEE TRANSACTIONS ON COMPUTATIONAL IMAGING,~Vol.~x, No.~x, August~xxxx}%
{Shell \MakeLowercase{\textit{et al.}}: Bare Demo of IEEEtran.cls for IEEE Journals}
%



\maketitle

\begin{abstract}
This paper presents an adaptive and intelligent sparse model for digital image sampling and recovery. In the proposed sampler, we adaptively determine the number of required samples for retrieving image based on space-frequency-gradient information content of image patches. By leveraging texture in space, sparsity locations in DCT domain, and directional decomposition of gradients, the sampler structure consists of a combination of uniform, random, and nonuniform sampling strategies. For reconstruction, we model the recovery problem as a two-state cellular automaton to iteratively restore image with scalable windows from generation to generation. We demonstrate the recovery algorithm quickly converges after a few generations for an image with arbitrary degree of texture. For a given number of measurements, extensive experiments on standard image-sets, infra-red, and mega-pixel range imaging devices show that the proposed measurement matrix considerably increases the overall recovery performance, or equivalently decreases the number of sampled pixels for a specific recovery quality compared to random sampling matrix and Gaussian linear combinations employed by the state-of-the-art compressive sensing methods. In practice, the proposed measurement-adaptive sampling/recovery framework includes various applications from intelligent compressive imaging-based acquisition devices to computer vision and graphics, and image processing technology. Simulation codes are available online for reproduction purposes.
\end{abstract}


%
\IEEEpeerreviewmaketitle

\section{Introduction}
%
%
%
%
\IEEEPARstart{C}{urrent} digital imaging devices at first acquire images and then separately compress them leading to big data-related problems. Contrarily, the aim of emerging Compressive Sensing (CS) theory is to merge sampling and compressing into one step by introducing the concept of ``compressing during sensing" for reducing data-rate, acquisition time, power consumption, and device manufacturing cost with demanding applications to next-generation Infra-Red (IR), remote sensing and Magnetic Resonance Imaging (MRI) systems \cite{eldar2012compressed, marvasti2012nonuniform, romberg2008imaging}. The theory of CS states that signals with a sparse representation in a specific domain can be recovered at a rate less than the traditional Shannon-Nyquist rate theorem \cite{donoho2006compressed, candes2006robust}. DCT, wavelet, and gradient spaces represent paradigms in which natural images are sparse or at least compressible. The later is not fully invertible.
%
%

\subsection{Considered Scenario and Related Arts}
In the literature, significant attempts have been done to compressively sample data at a very low sampling rate \cite{ji2008bayesian, shahrasbi2017model, wang2010variable, malloy2014near, yang2016high}. Conventional CS-based sampling/recovery strategies only exploit the sparsity-prior of signals in an appropriate domain \cite{tropp2007signal, blumensath2010normalized, marvasti2012sparse}. On the other hand, theoretical compressive sampling analyses rely on using Gaussian or random sampling matrices to measure signals \cite{candes2006near}. However, to design sensing matrix, recent scientific studies discover considering signal model-prior in sampling phase can significantly improve recovery quality \cite{shahrasbi2017model, wang2010variable, yang2016high}. In this regard, the present paper concentrates on finding a way to adaptively and sparsely sense image scene in order to simply and efficiently retrieve the image.

Ji et al. in \cite{ji2008bayesian} introduced an adaptive Bayesian CS (BCS) with abilities such as determining sufficient number of measurements and confidence of recovery. The researchers in \cite{shahrasbi2017model} proposed a nonuniform CS-based image sampling/recovery method in which a Hidden Markov Tree (HMT) is utilized to consider the correlation among sparse wavelet coefficients in both sampling and reconstructing phases, so-called uHMT-NCS. In \cite{wang2010variable}, a variable density sampling strategy was designed in the frequency domain which exploits priori statistical distributions of images in the wavelet space. Malloy and Nowak in \cite{malloy2014near} suggested an adaptive compressed sensing algorithm that in comparison to standard random nonadaptive design matrices requires nearly the same number of measurements but succeeds at lower SNR levels. Yang et al. in \cite{yang2016high} designed an efficient Mixed Adaptive-Random (MAR) sensing matrix which employs image edge information with a sensor array 16 times less than the ultimate length of recovered image. The Iterative Method with Adaptive Thresholding (IMAT) \cite{marvasti2012sparse} and its modified version equipped with interpolation, IMATI \cite{zayed2014new}, are extensions of the Iterative Hard Thresholding (IHT) family for sparse recovery \cite{blumensath2010normalized}. Instead of a fixed thresholding, they adaptively threshold coefficients in a transform domain such as Discrete Cosine Transform (DCT) during iterations. To recover sparse signals from Gaussian measurements, the authors in \cite{sadeghi2016iterative} presented a general iterative framework based on proximal methods called Iterative Sparsification-Projection (ISP). ISP family contains the well-known SL0 algorithm \cite{mohimani2009fast} as a special case. In \cite{duarte2009learning}, a learning-based approach by jointly optimizing the sensing and overcomplete dictionary matrices was introduced which performs better than random matrices or the matrices that are optimized without learning dictionary. In \cite{sadeghigol2016model}, a variational BCS in complex wavelet domain was suggested. This model, called TSCW-GDP-HMT, considers sparsity and structural information of images.


\subsection{Motivations and Contributions}
Although aforementioned efforts result in performance improvement somewhat, it seems that the role of Artificial Intelligence (AI) in the context of CS-based sampling/recovery is yet negligible. In this paper, we take a step towards considering AI and present an adaptive and intelligent model for digital image sampling and recovery. For a given number of measurements, we show the proposed sensing matrix considerably increases the overall recovery performance, or equivalently decreases the number of sampling points for a specific recovery quality compared to Gaussian, random, and dynamic sampling matrices employed by the state-of-the-art CS methods \cite{sadeghi2016iterative, yang2016high, sadeghigol2016model, ji2008bayesian, marvasti2012sparse, tropp2007signal}. The number of required samples in the spatial domain to reconstruct image is adaptively determined based on space-frequency-gradient information content of image. To this intent, a mixture of uniform, random, and nonuniform sampling strategies is suggested \cite{marvasti2012nonuniform}. The devised intelligent sampling mechanism donates the advantages of simple, fast, and efficient recovery compared to complicated methods \cite{shahrasbi2017model, sadeghi2016iterative, sadeghigol2016model}. For reconstruction, we model the recovery as a Cellular Automaton (CA) machine to iteratively restore the image with scalable windows from generation to generation. To the best of the authors' knowledge, this is the first work for modeling of image reconstruction issue via a CA. The convergence of the proposed CA-based recovery algorithm is guaranteed after a few generations, thus making it suitable for reconstructing the present mega-pixel range images, whereas sophisticated approaches such as ISP \cite{sadeghi2016iterative}, TSCW-GDP-HMT \cite{sadeghigol2016model}, and BCS \cite{ji2008bayesian} fail to process and store such high-dimensional signals with general-purpose computers.

The adaptivity in signal measurements helps to automatically select enough sampling points based on local information content of the scene under view. For instance, lowpass images are sampled with low rates, whereas cluttered scenes are acquired with more samples for an accurate reconstruction. Generally, available methods in the literature do not have such a degree of flexibility and for each signal are manually adjusted at a fixed rate. This undesirable phenomenon may lead to data-redundancy in smooth images or information-loss in textured ones. In practice, the measurement-adaptive image sampling/recovery scheme includes various applications from intelligent compressive imaging-based acquisition devices \cite{duarte2008single} to computer vision and graphics, and image processing technology, e.g. inpainting, remote sensing missing recovery, denoising of impulsive noise, and mesh-based image representation \cite{li2016anisotropic}. For reproduction purposes, simulation codes are available online\footnote{http://ee.sharif.edu/$\thicksim$imat/}.

Briefly, the main contributions of the paper are
\begin{itemize}
\item proposing a new measurement-adaptive image sampling mechanism consisting of three uniform, random, and nonuniform samplers which, in a synergistically manner, incorporates local space-frequency-gradient information content of the image,
\item introducing a novel nonuniform sampler based on directional gradients,
\item suggesting a novel cellular automaton model for image recovery which its convergence is guaranteed after a few generations, and
\item exposing various examinations to demonstrate the efficiency of the proposed sampling/recovery approach under different settings and application areas.
\end{itemize}

The remainder of this paper is organized as follows. Section~\ref{IntelligentSampler} describes the proposed sampler. In Section~\ref{CABasedRecoverer}, we explain the CA-based recoverer. Section~\ref{ExperimentalRes} evaluates the effectiveness of our framework and compares it to modern approaches. The paper is finally concluded in Section~\ref{ConclusionSection}.


\section{Intelligent Sampling}
\label{IntelligentSampler}
\begin{figure}[!t]
\centering
\includegraphics[width=\linewidth]{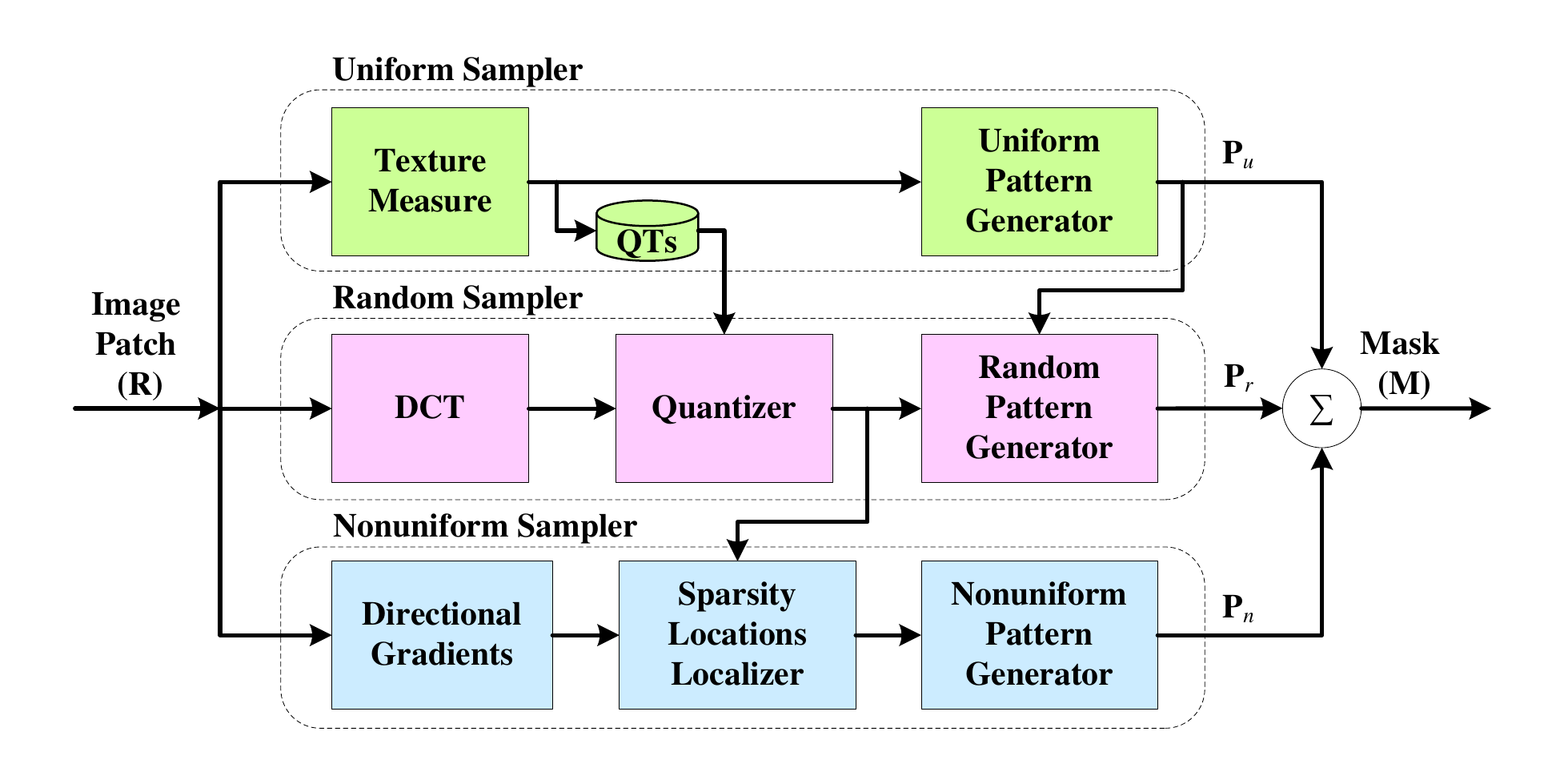}
\caption{The proposed measurement-adaptive sparse sampler.}
\label{SamplingScheme}
\end{figure}

Based on the content of local image patches, the proposed sampling method performs flexibly at various rates to capture required informative samples for recovery step. Figure~\ref{SamplingScheme} depicts the block diagram of the suggested measurement-adaptive sampler. As shown, the system gets a gray-level image patch and gives the related binary mask. Due to psycho-visual considerations, we assume that patches are square with the length $b=8$. The final sampling mask is generated from the union of three uniform, random, and nonuniform patterns obtained from spatial, frequency, and gradient spaces, respectively. In the further sub-sections, we discuss about each sampling strategy in detail.

\subsection{Texture Measure}
To adaptively sense the information content of each local image patch, it may be required to measure a representative criterion of signal's complexity \cite{taimori2015quantization}. Here, we utilized Shannon's joint entropy determined from Haralick's co-occurrence matrix \cite{haralick1973textural, jain1995machine} as a quantitative metric to measure image block texture. We calculate the texture percentage $\eta\in \left[0,\ 100\right]$ as
\begin{equation}
\label{localtexture}
\eta=\frac{100}{H_{{\rm max}}}\sum^B_{i=1}{\sum^B_{j=1}{{\hat{p}}_{i,j}{{\log }_2 \left(\frac{1}{{\hat{p}}_{i,j}}\right)\ }}},
\end{equation}
where the gray-level co-occurrence matrix $\widehat{{\mathbf P}}={\left[{\hat{p}}_{i,j}\right]}_{B\times B}$ is the joint Probability Mass Function (PMF) which is obtained by the quantized or scaled version of the input image patch, namely ${{\mathbf G}}_s={\left[g^s_{i,j}\right]}_{b\times b}$. The joint probabilities can be defined in the horizontal, vertical, main diagonal, and off-diagonal directions. We used the horizontal neighbor as defined in MATLAB by ${\hat{p}}_{i,j}=P\left(g^s_{r,c}=i,g^s_{r,c+1}=j\right)$, where $1\le i,j\le B$, $1\le r\le b$, and $1\le c\le b-1$. The variables $r$ and $c$ represent row and column, respectively. We considered the number of distinct gray levels in ${{\mathbf G}}_s$ as $B=8$, hence $g^s_{i,j}\in \left[1,\ B\right],\ 1\le i,j\le b$. The normalizer $H_{{\rm max}}$ in \eqref{localtexture} denotes the maximum entropy of $\widehat{{\mathbf P}}$, which obtains from the uniform distribution, i.e. ${\hat{p}}_{i,j}=\frac{1}{B^2}$, $\forall$ $1\le i,j\le B$. Thus, $H_{{\rm max}}=\sum^B_{i=1}\sum^B_{j=1}\frac{1}{B^2}{\log }_2 (B^2)$, which $B=8\Rightarrow H_{{\rm max}}=6$.

\subsection{Uniform Sampler}
\begin{figure}[!t]
\centering
\includegraphics[width=6cm]{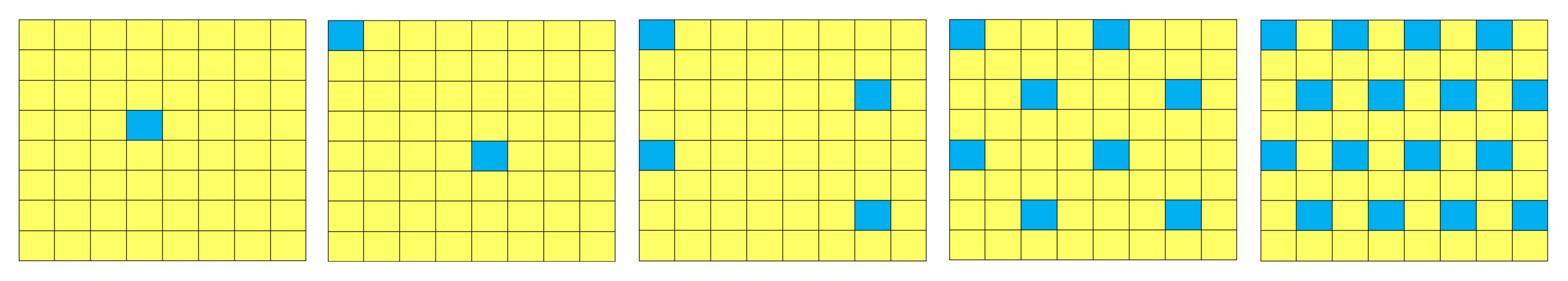}
\caption{From left to right, patterns of the uniform sampler defined by matrices ${{\mathbf P}}_{{\rm UVL}}$, ${{\mathbf P}}_{{\rm ULT}}$, ${{\mathbf P}}_{{\rm UBT}}$, ${{\mathbf P}}_{{\rm UHT}}$, and ${{\mathbf P}}_{{\rm UVH}}$.}
\label{UniformPatterns}
\end{figure}

In the proposed uniform sampling strategy, we punch local patches at certain regular locations based on their local texture content in space domain. In fact, the punch operation makes appropriate regular patterns for a stable recovery. As shown in Fig.~\ref{SamplingScheme}, the uniform sampler at first gets the image patch ${\mathbf R}={[r_{i,j}]}_{b\times b\ }$. Afterwards, the textureness of the signal is estimated. The number of punched points is proportional to the calculated texture percentage. By this intuition, whenever the textureness is high, we decorate the image with denser regular points and vice versa. Therefore, the binary mask of uniform pattern is determined by the following fuzzy rule

\begin{equation}
\label{UniformPatt}
{{\mathbf P}}_u=\left\{ \begin{array}{cc}
{{\mathbf P}}_{{\rm UVL}}, & 0\le \eta<10 \\
{{\mathbf P}}_{{\rm ULT}}, & 10\le \eta<25 \\
{{\mathbf P}}_{{\rm UBT}}, & 25\le \eta<45 \\
{{\mathbf P}}_{{\rm UHT}}, & 45\le \eta<70 \\
{{\mathbf P}}_{{\rm UVH}}, & 70\le \eta\le 100 \end{array}
\right.,
\end{equation}
in which the matrices ${{\mathbf P}}_{{\rm UVL}}$, ${{\mathbf P}}_{{\rm ULT}}$, ${{\mathbf P}}_{{\rm UBT}}$, ${{\mathbf P}}_{{\rm UHT}}$, and ${{\mathbf P}}_{{\rm UVH}}$ are assigned for very-low, low, bandpass, high, and very-high textures, respectively. Figure~\ref{UniformPatterns} shows the regular lattices employed in the proposed uniform sampler. The number of live cells (punched points) in the matrices ${{\mathbf P}}_{{\rm UVL}}$, ${{\mathbf P}}_{{\rm ULT}}$, ${{\mathbf P}}_{{\rm UBT}}$, ${{\mathbf P}}_{{\rm UHT}}$, and ${{\mathbf P}}_{{\rm UVH}}$ are $2^0$, $2^1$, $2^2$, $2^3$, and $2^4$, respectively. This reveals the relation between the texture and the number of live cells is nonlinear with an increasing exponential form. The configuration of punched points affects the recovery performance. Sub-optimal lattices should have minimum number of sample points and maximum number of live cells in a predefined neighboring of missing samples. They should also consider patch boundaries and local correlation. We tried to experimentally design patterns based on such constraints.

\subsection{Random Sampler}
The intermediate layer of Fig.~\ref{SamplingScheme} depicts the proposed random sampler. Here, we first convert the data matrix ${\mathbf R}$ from the spatial domain to the frequency space, ${{\mathbf R}}_f \in {{\mathbb R}}^{b\times b}$, by using 2-D DCT transform as ${{\mathbf R}}_f={\mathbf T}{{\mathbf R}}{{\mathbf T}}^{{\rm T}}$, where the matrix $\mathbf T$ represents the transformation kernel of DCT. In DCT domain, samples are decorrelated and sparse. After an adaptive quantization of DCT coefficients, we estimate the rate of random sampling. Accordingly, a uniformly distributed random mask is finally generated. In the following sections, we discuss about creating this mask.

\subsubsection{Random Sampling Rate Measure}
To measure sparsity, one way is to threshold insignificant coefficients in a transform domain \cite{marvasti2012sparse}. However, such thresholding may be inaccurate due to varying information of the signal under process/acquisition. To overcome this problem, we suggest adaptive quantization of DCT coefficients inspired from the following observation.

\newtheorem{Observation1}{Observation}
\begin{Observation1}[Adaptive quantization]
\label{Obser1}
  In JPEG compression standard where DCT transform is utilized, the results in \cite{taimori2015quantization} reveal that the number of zero entries of quantized DCT coefficients is inversely proportional to both compression quality level, $l$, and the image block texture, $\eta$. Now, let the matrix $\widetilde{\mathbf R}_{f}=[{\tilde{r}}^{f}_{i,j}] \in {{\mathbb Z}}^{b\times b}$ be the quantized DCT matrix for a $b \times b$ image patch, then ${\parallel \widetilde{\mathbf R}_{f} \parallel}_{\ell_0} = b^2- \mathcal{Z}(\widetilde{\mathbf R}_{f})$, where ${\parallel \cdot \parallel}_{\ell_p}$ denotes the ${\ell_p}$-norm and the operator $\mathcal{Z}(\cdot)$ counts the number of zero entries of a matrix/vector. From the above results, we have both ${\parallel \widetilde{\mathbf R}_{f} \parallel}_{\ell_0} \propto \frac{1}{l}$ and ${\parallel \widetilde{\mathbf R}_{f} \parallel}_{\ell_0} \propto \frac{1}{\eta}$. By assuming their equivalence, we have $\eta\propto l$. This shows the texture percentage defined in \eqref{localtexture} can be a good criterion for adaptive quantization of information in the frequency domain to measure the necessary random sampling rate.
  \end{Observation1}

  \emph{Details:} Based on Observation 1, we calculate the quantized DCT matrix as
  \begin{equation}
  \label{QuantizedDCTMatrix}
  \widetilde{\mathbf R}_{f}=\lfloor {\mathbf R}_{f}\odiv {{\mathbf Q}}_\eta \rceil,
  \end{equation}
where the symbols $\odiv $ and $\left\lfloor \cdot \rceil \right.$ denote the entry-by-entry division and the nearest integer, respectively. The matrix ${{\mathbf Q}}_\eta={\left[q^{\eta}_{i,j}\right]}_{b\times b}$, $\forall q^{\eta}_{i,j}\in {{\mathbb N}}$, also represents the quantization table related to the texture $\eta$. For determining ${{\mathbf Q}}_\eta$, we exploited the Quantization Tables (QTs) introduced by Independent JPEG Group (IJG) \cite{luo2010jpeg}. Based on the value of $\eta$, the corresponding quantization matrix is chosen and applied (See Appendix.). Therefore, we adaptively derive sparsity number from the number of nonzero entries of quantized DCT coefficients by $k={\parallel \widetilde{\mathbf R}_{f} \parallel}_{\ell_0}$. Now, we suggest measuring the rate of random sampling as the following formula, which is consistent with the theory of CS, i.e. $c\cdot k\cdot \log_{10} (\frac{n}{k})$,
\begin{equation}
\label{RandomSamplingRate}
{R_{\rm rs}}=\left\{ \begin{array}{cc}
\lfloor c\cdot k\cdot \log_{10} (d\cdot \frac{n}{k}) \rceil, & k\neq 0  \\
0, & \rm otherwise \end{array}
\right.,
\end{equation}
in which $n=b^2$, and $c$ and $d$ are tunable factors. The parameter $c$ changes the number of undersampled points, whereas the coefficient $d$ prevents from decaying sampling rate on well-textured patches and saturates it at a relatively fixed rate to control unnecessary information. By setting $c=1.3$ and $d=2.8$, the function in \eqref{RandomSamplingRate} is a non-decreasing curve vs $k$. It is important to note that, in practical situations, the number of non-zero quantized DCT coefficients, $k$, may be a large value even near the length of signal, $n$. This phenomenon occurs especially in high quality level settings or textured regions. Generally, high performance compressive sensing in such a dense scenario is problematic because compressive sensing theory for real-world applications imposes the sampling rate to be at most 10\% \cite{lustig2008compressed}. We tried to consider this limitation for calculating the rate of random sampling in \eqref{RandomSamplingRate} by setting $c=1.3$ and $d=2.8$, which covers both sparse and dense cases.


\subsubsection{Graduate Randomization Procedure}
Instead of blind random sampling adopted by researchers of compressive sensing field, we introduce a Graduate Randomization Procedure (GRP). In this procedure, by receiving both the feedback of the sampling rate in frequency domain and the generated uniform mask, we add a certain randomness to the uniform pattern. In other words, from low- to high- textured patches, GRP gradually adds random nonuniformity to the uniform base lattice. It is important to note that random samples are selected and added to locations other than uniform marks. In order to generate the random pattern, the GRP algorithm is implemented as follows
\begin{itemize}
\item Find the number of zero elements in the uniform pattern, i.e. $\mathcal{Z}({{\mathbf P}}_u)$, and their locations by storing in the set ${\mathcal L}\triangleq {\left\{(p^i, q^i)\right\}}^{\mathcal{Z}({{\mathbf P}}_u)}_{i=1}$, where $p^i$ and $q^i$ represent the row and column of the $i^{\textnormal{th}}$ location, respectively.
\item Shuffle the location of zero entries and substitute the set $\mathbf {\mathcal L}$ with the result.
\item Truncate \eqref{RandomSamplingRate} by
\begin{equation}
\label{TruncatedRandomSamplingRate}
{R_{\rm rs}}=\left\{ \begin{array}{cc}
\mathcal{Z}({\mathbf P}_u), & R_{\rm rs} > \mathcal{Z}({\mathbf P}_u)  \\
R_{\rm rs}, & \rm otherwise \end{array}
\right..
\end{equation}
\item Generate $R_{\rm rs}$ integer values randomly without replacement in the range $[1, \mathcal{Z}({\mathbf P}_u)]$, namely $\mathbf v = [v_1, v_2, \cdots, v_{R_{\rm rs}}]^{\rm T}$.
\item Define the matrix of random pattern, $\mathbf P_r={[p^{r}_{i,j}]}_{b\times b\ }$, and initialize it with $\mathbf P_r =\mathbf O$, where $\mathbf O$ represents a ${b\times b\ }$ zero matrix.
\item For $n=1$ to $R_{\rm rs}$, at first, assign $(i, j)\leftarrow {\mathcal L}\{v_n\}$, and then $p^{r}_{i,j}\leftarrow 1$.
\end{itemize}

\subsection{Nonuniform Sampler}
\begin{figure}[!t]
\centering
\includegraphics[width=2.5cm]{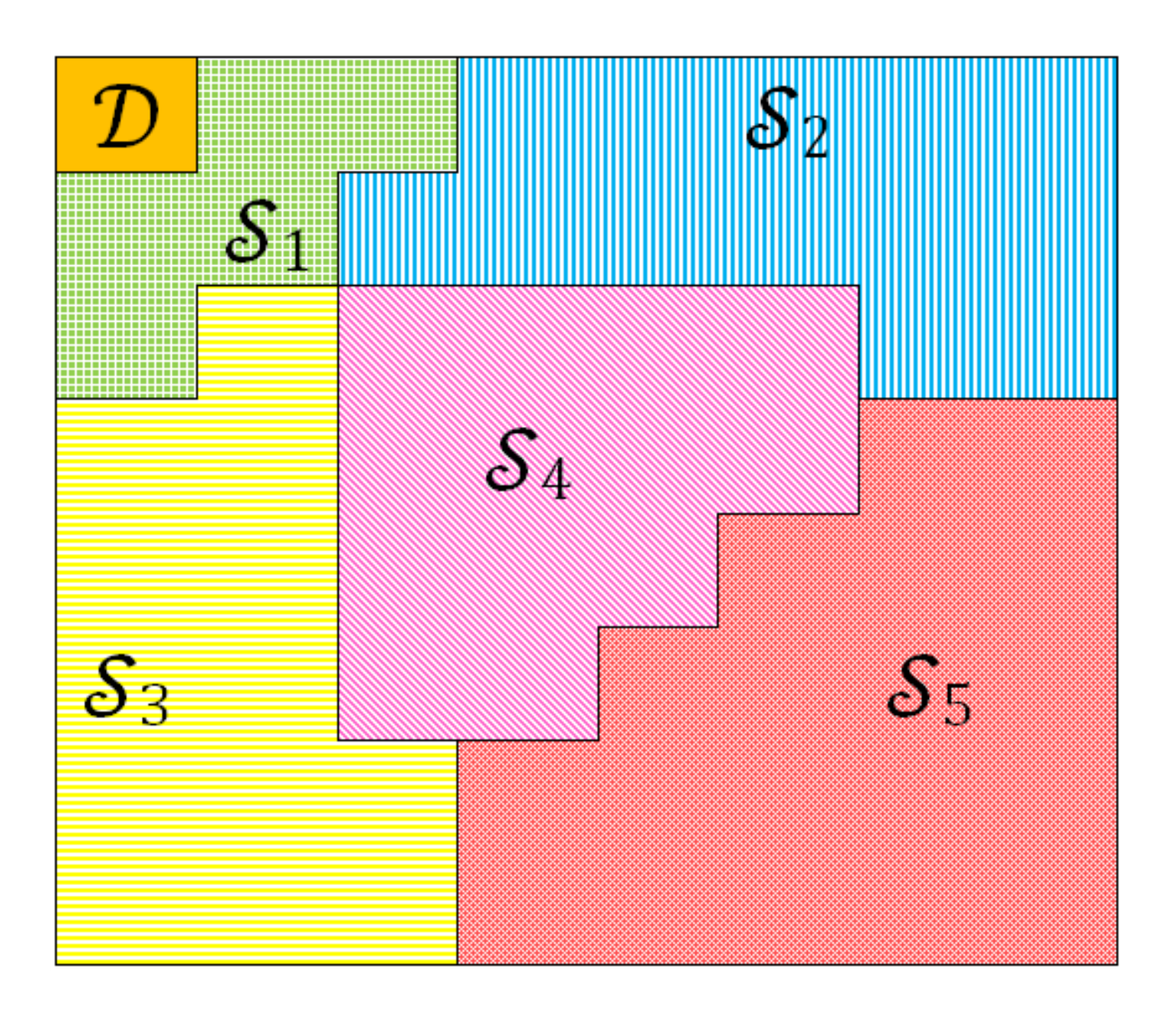}
\caption{A partitioning of 2-D frequency domain defined in \cite{hasimoto2004transform, taimori2015quantization} into 6 regions.}
\label{ModePartitioning} 
\end{figure}
It is very important to preserve the structure of image like edges in recovery side. Generally, edge preservation is a difficult task in interpolation even for sophisticated algorithms. Due to fuzziness nature of image edges, an edge-prior sampler can reduce uncertainty in reconstruction phase. Although, the GRP carry this task somewhat, in well-structured horizontal, vertical, and diagonal edges, or periodic configurations which their recovery is perceptually important from subjective evaluation viewpoint, more edge samples are required to reduce uncertainty and unambiguously recover the missing scattered samples. Hence, we suggest a sparsity location-aware algorithm to capture horizontal, vertical, and diagonal edges in sampling phase.

To implement the above idea, we address sparsity locations in different frequency regions. Figure~\ref{ModePartitioning} depicts a partitioning of 2-D frequency domain defined in \cite{hasimoto2004transform, taimori2015quantization}. Except for the single DC mode ${\mathcal D}$, the five subsets of AC modes exist, namely ${\mathcal S_1}$ to ${\mathcal S_5}$ corresponding to low, horizontal, vertical, diagonal, and high frequencies information, respectively. If nonzero quantized DCT coefficients are sparsely located in horizontal, vertical, or diagonal regions, the corresponding edge points are sampled in space domain. For instance, in a block of well-structured vertical edges, the nonzero coefficients are only appeared in the partitions ${\mathcal S_1}$ and ${\mathcal S_2}$. To detect the edge maps, we first determine a $D$-directional decomposition of the image patch gradient \cite{bai2005study, huang2003gradient}. For this purpose, we utilized Sobel operator to calculate the gradient of image block, $\mathbf{g}\triangleq [\frac{\partial r_{i,j}}{\partial x}, \frac{\partial r_{i,j}}{\partial y}]^{\rm T}$, $\forall i,j$, and considered $D=8$ with four cardinal directions N, E, S, W in addition to four ordinal directions NE, SE, SW, NW to decompose the block. In order to preserve the image structure in the boundary of blocks, we also acquired border pixels for computing the gradient. Let $\mathbf{F}_{d_\gamma}=[f^{d_\gamma}_{i,j}]_{b\times b}$, $\forall \gamma \in [1, D]\equiv \{\rm N, NW, W, SW, S, SE, E, NE\}\equiv \{\emph{d}_1, \emph{d}_2, \emph{d}_3, \emph{d}_4, \emph{d}_5, \emph{d}_6, \emph{d}_7, \emph{d}_8\}$, be $D$-directional gradient features. To project the gradient vector $\mathbf{g}$ in $(x$-$y)$ plane onto the two adjacent directions
\begin{equation}
\label{Alpha1}
\alpha_1=\left\{ \begin{array}{cc}
\frac{|(|\frac{\partial r_{i,j}}{\partial x}|-|\frac{\partial r_{i,j}}{\partial y}|)|}{{\parallel \mathbf{g} \parallel}_{\ell_2}}, & {\parallel \mathbf{g} \parallel}_{\ell_2}\neq 0  \\
0, & \rm otherwise \end{array}
\right.,
\end{equation}
\begin{equation}
\label{Alpha2}
\alpha_2=\left\{ \begin{array}{cc}
\frac{\sqrt{2} \min(|\frac{\partial r_{i,j}}{\partial x}|, |\frac{\partial r_{i,j}}{\partial y}|)}{{\parallel \mathbf{g} \parallel}_{\ell_2}}, & {\parallel \mathbf{g} \parallel}_{\ell_2}\neq 0  \\
0, & \rm otherwise \end{array}
\right.,
\end{equation}
the following rules are utilized.
\begin{itemize}
  \item If $\frac{\partial r_{i,j}}{\partial x}\geq 0 \wedge |\frac{\partial r_{i,j}}{\partial y}| \leq |\frac{\partial r_{i,j}}{\partial x}|$, then assign $f^{d_1}_{i,j}\leftarrow \alpha_1$; else if $\frac{\partial r_{i,j}}{\partial x}< 0 \wedge |\frac{\partial r_{i,j}}{\partial y}| \leq |\frac{\partial r_{i,j}}{\partial x}|$, then $f^{d_5}_{i,j}\leftarrow \alpha_1$; else if $\frac{\partial r_{i,j}}{\partial x}\geq 0 \wedge |\frac{\partial r_{i,j}}{\partial y}| > |\frac{\partial r_{i,j}}{\partial x}|$, then $f^{d_7}_{i,j}\leftarrow \alpha_1$; else if $\frac{\partial r_{i,j}}{\partial x}< 0 \wedge |\frac{\partial r_{i,j}}{\partial y}| > |\frac{\partial r_{i,j}}{\partial x}|$, then $f^{d_3}_{i,j}\leftarrow \alpha_1$.
  \item If $\frac{\partial r_{i,j}}{\partial x}\geq 0 \wedge \frac{\partial r_{i,j}}{\partial y}\geq 0$, then assign $f^{d_8}_{i,j}\leftarrow \alpha_2$; else if $\frac{\partial r_{i,j}}{\partial x}\geq 0 \wedge \frac{\partial r_{i,j}}{\partial y}< 0$, then $f^{d_2}_{i,j}\leftarrow \alpha_2$; else if $\frac{\partial r_{i,j}}{\partial x}< 0 \wedge \frac{\partial r_{i,j}}{\partial y}< 0$, then $f^{d_4}_{i,j}\leftarrow \alpha_2$; else if $\frac{\partial r_{i,j}}{\partial x}< 0 \wedge \frac{\partial r_{i,j}}{\partial y}\geq 0$, then $f^{d_6}_{i,j}\leftarrow \alpha_2$.
\end{itemize}

\begin{figure}[!t]
\centering
\includegraphics[width=4cm]{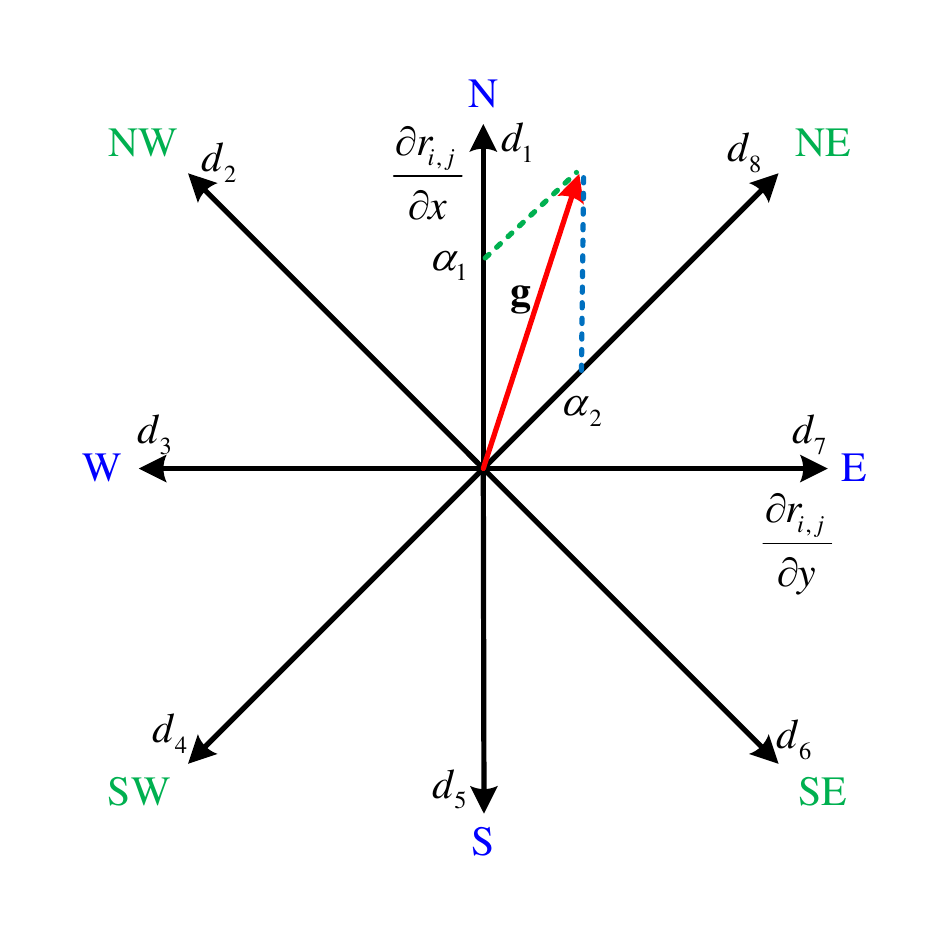}
\caption{The gradient directions corresponding to cardinal and ordinal directions, and the directional decomposition of the gradient vector $\mathbf{g}$ into two $d_1$ and $d_8$ adjacent directions.}
\label{8-DirectionalGradient} 
\end{figure}

Afterwards, we apply a threshold to the magnitude of two directions with stronger gradients to suppress redundant uninformative samples. Figure~\ref{8-DirectionalGradient} illustrates the directional decomposition of the gradient vector $\mathbf{g}$ into two adjacent directions. For this example, we have $f^{d_1}_{i,j}= \alpha_1$, $f^{d_2}_{i,j}= 0$, $f^{d_3}_{i,j}= 0$, $f^{d_4}_{i,j}= 0$, $f^{d_5}_{i,j}= 0$, $f^{d_6}_{i,j}= 0$, $f^{d_7}_{i,j}= 0$, and $f^{d_8}_{i,j}= \alpha_2$ based on the specified rules. The steps of nonuniform sampling algorithm are implemented as follows.
\begin{itemize}
  \item Initialize $D$-directional gradient matrices, i.e. $\mathbf{F}_{d_\gamma}=\mathbf{O}$, $\forall \gamma \in [1, D]$.
  \item Extract $D$-directional gradient features as explained above for the vector $\mathbf{g}$.
  \item Obtain the normalized versions of $\mathbf{F}_{d_\gamma}$, $\forall \gamma \in [1, D]$, between $0$ and $1$.
  \item Define the matrix of nonuniform pattern, $\mathbf P_n={[p^n_{i,j}]}_{b\times b\ }$, and initialize it as $\mathbf P_n =\mathbf O$.
  \item For determining the structure of edges, initialize a binary string with the length of the number of AC frequency regions, $N_{\rm fr}=5$, such as $\mathbf{a}=``a_1a_2a_3a_4a_5"=``00000"$.
  \item For $s=1$ to $N_{\rm fr}$, obtain the set ${\mathcal{B}_s}\triangleq \{\tilde{r}^{f}_{i,j}|(i,j)\in \mathcal{S}_s\}$. If ${\parallel {\mathcal{B}_s} \parallel}_{\ell_0}\neq 0$, set $a_s \leftarrow ``1"$.
  \item If the string $\mathbf{a}=``11000"\vee ``10100"\vee ``10010"\vee ``11010"\vee ``10110"$, which respectively represent vertical, horizontal, diagonal, vertical-diagonal, or horizontal-diagonal edges, at first, sort the set $\{\sum^b_{i=1} \sum^b_{j=1} f^{d_\gamma}_{i,j}\}^D_{\gamma=1}$. Afterwards, for the two directions having stronger gradients, calculate its gradient magnitude, and then renormalize it between $0$ and $1$. Finally, for $i=1$ to $b$ and $j=1$ to $b$, if the resulting magnitude of gradient exceeds the predefined threshold $0 \leq \tau \leq 1$, set $p^n_{i,j}\leftarrow 1$.
\end{itemize}

\subsection{Measurement-Adaptive Sampling Algorithm}
As shown in Fig.~\ref{SamplingScheme}, the ultimate binary mask is generated by the union of uniform, random, and nonuniform patterns as
\begin{equation}
\label{FinalMask}
\mathbf{M}= \bigcup_{i\in \{u,r,n\}}{\mathbf{P}_i}.
\end{equation}
The regularity, randomness, and structure of patterns in the above combined form create naturally a sampling mask with chaotic behavior. For selecting the threshold $\tau$ in the nonuniform sampler, a trade-off between subjective quality enhancement and increased redundancy exists. If $\tau \rightarrow 1$, then $\mathbf{M}\rightarrow {\mathbf{P}_u}\bigcup {\mathbf{P}_r}$. Therefore, we empirically set $\tau=0.9$ to maintain sampling near an optimal sub-Shannon-Nyquist rate. The downsampled image patch is constructed from
\begin{equation}
\label{SampledPatch}
\mathbf{S} = \mathbf{M} \odot \mathbf{R},
\end{equation}
in which the symbol $\odot$ denotes Hadamard product. Algorithm~\ref{SamplingAlgorithm} summarizes the proposed adaptive intelligent image sampling strategy. For the previously captured image $\mathbf{I}$ of the dimension $h\times w$, the proposed block-wise algorithm can be repeated on all nonoverlapping image patches to get the overall sampling mask ${\mathbf M}_s=[m^s_{i,j}]_{h\times w}$ and the subsampled image ${\mathbf I}_s=[i^s_{i,j}]_{h\times w}$.

\begin{algorithm}[!t]
\caption{The proposed adaptive sampling algorithm}
\label{SamplingAlgorithm}
\begin{algorithmic}[1]
\State \textbf{Input}: The image patch ${\mathbf R}$.
\State Measure the texture $\eta$ as \eqref{localtexture}.
\State Generate the uniform sampling mask $\mathbf{P}_u$ by \eqref{UniformPatterns}.
\State Determine the quantization table $\mathbf{Q}_\eta$ in terms of $\eta$.
\State Transform the patch ${\mathbf R}$ to 2-D DCT domain ${\mathbf R}_f$.
\State Quantize DCT coefficients via $\widetilde{\mathbf R}_{f}=\lfloor {\mathbf R}_{f}\odiv {{\mathbf Q}}_\eta \rceil$.
\State Estimate the sparsity number $k={\parallel \widetilde{\mathbf R}_{f} \parallel}_{\ell_0}$.
\State Calculate the rate of random sampling as \eqref{RandomSamplingRate}.
\State Perform the GRP routine to obtain the random mask $\mathbf{P}_r$.
\State Run the nonuniform sampler algorithm to generate $\mathbf{P}_n$.
\State Combine the created masks as $\mathbf{M}= \bigcup_{i\in \{u,r,n\}}{\mathbf{P}_i}$.
\State Obtain the scattered sample points by $\mathbf{S} = \mathbf{M} \odot \mathbf{R}$.
\State \textbf{Outputs}: The mask ${\mathbf M}$ and the subsampled image ${\mathbf S}$.
\end{algorithmic}
\end{algorithm}

\section{Cellular Automaton-based Image Recovery}
\label{CABasedRecoverer}
\begin{figure}[!t]
\centering
\includegraphics[width=7cm]{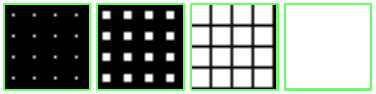}
\caption{Applying Algorithm~\ref{RecoveryAlgorithm} on a $32\times32$ DC image. From left to right, sampling matrices encompassed by the green-colored frame illustrate initial state, and the evolution at $1^{\rm st}$, $2^{\rm nd}$, and $3^{\rm rd}$ generations, respectively.}
\label{Convergence}
\end{figure}
As mentioned in the literature, different approaches exist for image recovery \cite{shahrasbi2017model, wang2010variable, malloy2014near, duarte2009learning, sadeghi2016iterative, marvasti2012sparse, sadeghigol2016model, blumensath2010normalized, tropp2007signal, ji2008bayesian, mohimani2009fast}. Here, we tailor a novel technique to recover scattered samples obtained from the intelligent sampling stage. We model the reconstruction mechanism by a dynamic CA. Cellular automata are simple intelligent agents with fundamental characteristics of locality, uniformity, and synchronicity \cite{schiff2011cellular}. They are composed of discrete cells equipped with special rules to be able to solve sophisticated problems. Various applications have been found for CA in image processing tasks such as noise removal, edge detection, and forgery localization \cite{rosin2014cellular}.

\subsection{Cellular Automaton Modeling for Recovery}
In modeling, we consider a 2-state CA machine, in which the states of dead and live cells represent the zero and one values in the sampling mask ${\mathbf M}_s$, respectively. Contrary to conventional fixed-neighbor CA models, the proposed CA-based recoverer performs as an iterative method which applies variable-scale windows to the sampled image ${\mathbf I}_s$. The size of square window, $\Omega$, increases at each generation of CA. In summary, we define the elements of model as $\mathcal{M}({\mathbf M}_s, {\mathbf I}_s, \Omega, \sigma, \zeta, \widehat{\mathbf I}, {\mathbf M}_g)$, in which $\sigma$ represents the standard deviation of a Gaussian kernel, $\zeta\geq1$ is a coefficient for increasing $\sigma$ during generations, and ${\mathbf M}_g$ and $\widehat{\mathbf I}$ denote the matrix at next generation of ${\mathbf M}_s$ and the recovered image, respectively. The rule behind the suggested recovery algorithm is simple yet efficient as follows. By considering the correlation among adjacent samples, missing pixels are reconstructed via a Gaussian-weighted averaging. We find live cells of ${\mathbf I}_s$ around the central dead cell $m^s_{i,j}$ and extract their corresponding weights in $(\Omega-1)$-Moore neighborhood \cite{schiff2011cellular}, namely the dynamic-length vectors $\mathbf{x}=[x_1,\cdots,x_k]^{\rm T}$ and ${\boldsymbol \omega}=[\omega_1,\cdots,\omega_k]^{\rm T}$, respectively. If at least one live cell exists, then, the dead cell $m^s_{i,j}$ is replaced with the weighted mean of live cells in the subsampled image and that dead cell will revive at next generation. At each generation of CA-based recoverer, the model elements are updated. Algorithm~\ref{RecoveryAlgorithm} presents the pseudo-code of CA-based image recovery algorithm, in which the symbols $\lfloor\cdot\rfloor$ and $\lceil\cdot\rceil$ represent the floor and ceiling functions, respectively. We utilized the replication rule to cope with boundary conditions of border cells.

After the last generation, we utilize a post processing stage to alleviate possible blockiness artifacts due to the patch-based nature of recovery. This phenomenon may be observed in plain regions. Based on the coding style of measurement-adaptive sampler, the number of live cells within a window can be an appropriate criterion to discriminate plain regions from textured ones in the recovery phase. Therefore, we apply a conditional smoothing filter only for flat regions. To this intent, at first, we initialize the post-processed result $\widehat{\mathbf {I}}_p=[\widehat{i}^{p}_{i,j}]_{h\times w}$ as $\widehat{\mathbf {I}}_p=\widehat{\mathbf I}$. Afterwards, for all $i\in[1,h]$ and $j\in[1,w]$, we scan the image $\widehat{\mathbf I}=[\widehat{i}_{i,j}]_{h\times w}$ with a square window of the length $\Omega_f=3$. If the number of live cells in the initial sampling mask $\mathbf{M}^{(0)}_s$ inside the window are less than the threshold $\lfloor \rho\cdot\Omega^2_f\rceil$, the corresponding pixel values of $\widehat{\mathbf I}$ in the window are smoothed via the Gaussian kernel $\mathbf{W}_{\sigma_f}=[w^{\sigma_f}_{i,j}]_{\Omega_f\times \Omega_f}$, for which $0\leq\rho\leq 1$. Then, the result is placed into $\widehat{i}^{p}_{i,j}$. In experiments, we set the coefficient $\rho=0.3$ and the standard deviation of smoother $\sigma_f=1.5$.

\begin{algorithm}[!t]
\caption{The proposed CA-based recovery algorithm}
\label{RecoveryAlgorithm}
\begin{algorithmic}[1]
\State \textbf{Inputs}: The mask ${\mathbf M}_s$ and the subsampled image ${\mathbf I}_s$.
\State Initialize $\Omega=3$, $\sigma=1$, $\zeta=1.05$, $\widehat{\mathbf I}={\mathbf I}_s$, ${\mathbf M}_g={\mathbf M}_s$, and $\mathbf{M}^{(0)}_s=\mathbf{M}_s$.
\State Calculate the number of dead cells by $N_d=\mathcal{Z}({\mathbf M}_g)$.
\While {$N_d\neq 0$}
   \State $\omega_h\triangleq\lfloor \frac{\Omega}{2}\rfloor$
   \State Obtain the Gaussian kernel $\mathbf{W_\sigma}=[w^{\sigma}_{i,j}]_{\Omega\times \Omega}$.
   \For {$r \leftarrow 1, h$}
        \For {$c \leftarrow 1, w$}
            \If{$m^s_{r,c}=0$}
                \State ${\vartheta_{\rm flag}}\triangleq0$
                \State $t\triangleq1$
                \For {$p \leftarrow -\omega_h, \omega_h$}
                     \For {$q \leftarrow -\omega_h, \omega_h$}
                         \If{$m^s_{r+p,c+q}=1$}
                            \State $x_t\triangleq i^s_{r+p,c+q}$
                            \State $\omega_t\triangleq w^{\sigma}_{{p+\lceil \frac{\Omega}{2}\rceil},{q+\lceil \frac{\Omega}{2}\rceil}}$
                            \State $\vartheta_{\rm flag}= 1$
                            \State $t \leftarrow t+1$
                         \EndIf
                     \EndFor
                \EndFor
                \If{$\vartheta_{\rm flag}=1$}
                   \State $\hat{i}_{r,c}= \frac{\mathbf{x}^{\rm T}{\boldsymbol \omega}}{\Sigma_k\omega_k}$
                   \State $m^g_{r,c}= 1$
                \EndIf
                \State Release the vectors $\mathbf{x}$ and ${\boldsymbol \omega}$.
            \EndIf
        \EndFor
    \EndFor
    \State $\Omega\leftarrow \Omega+2$
    \State $\sigma\leftarrow \zeta\cdot\sigma$
    \State ${\mathbf I_s}= \widehat{\mathbf I}$
    \State $\mathbf M_s= \mathbf M_g$
    \State $N_d=\mathcal{Z}({\mathbf M}_g)$
\EndWhile
\State Apply the post processing on $\widehat{\mathbf I}$ via $\mathbf{M}^{(0)}_s$ to obtain $\widehat{\mathbf {I}}_p$.
\State \textbf{Output}: The recovered post-processed image $\widehat{\mathbf {I}}_p$.
\end{algorithmic}
\end{algorithm}

\begin{figure}[!t]
\centering
\includegraphics[width=\linewidth]{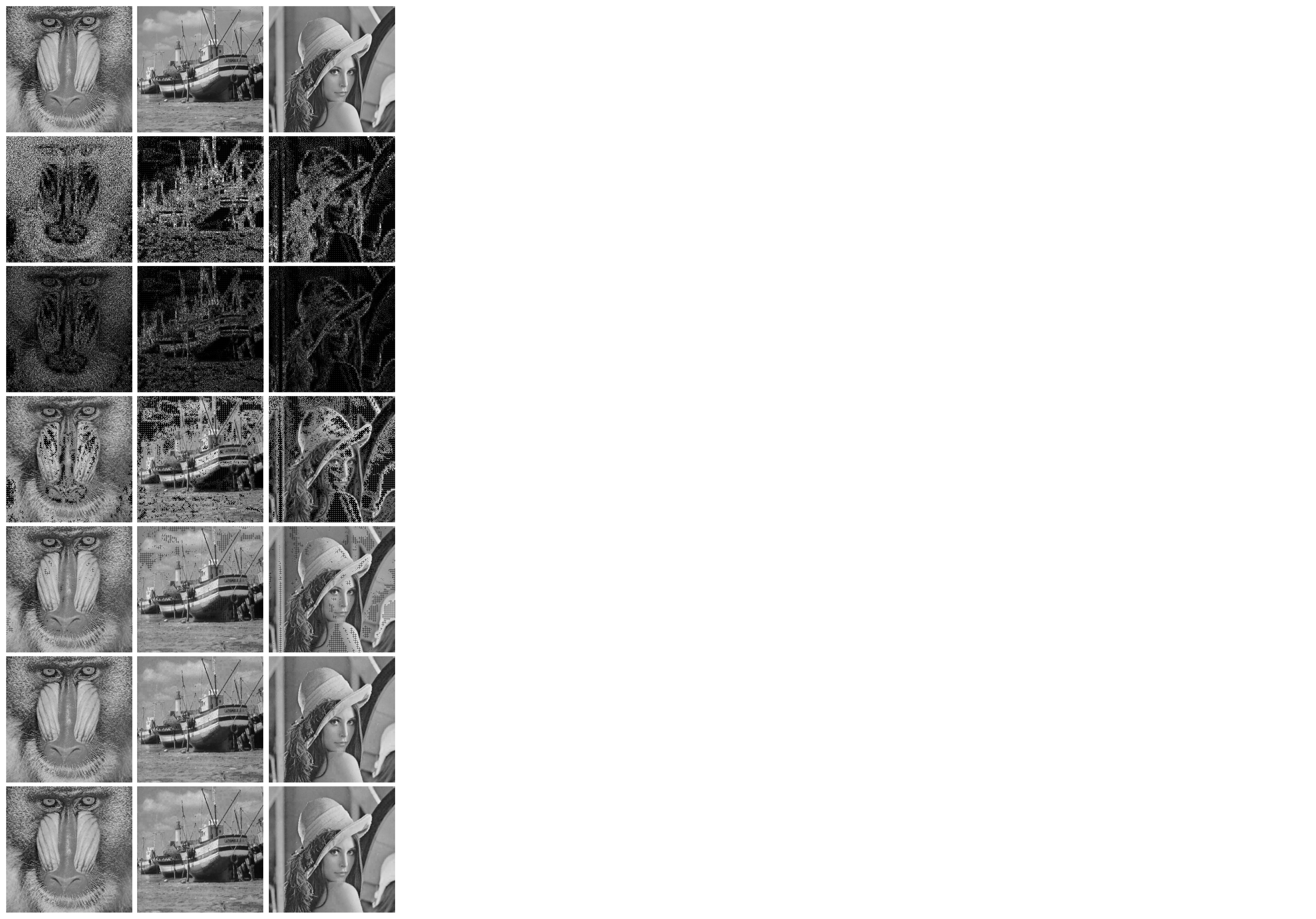}
\caption{From top to down, different images respectively show the original images, sampling masks, subsampled images, recovered results after $1^{\rm st}$, $2^{\rm nd}$ and $3^{\rm rd}$ generations, and the results of applying conditional smoother.}
\label{PipelineExample}
\end{figure}

\subsection{Convergence Analysis}
As illustrated in Fig.~\ref{PipelineExample}, the CA-based recoverer first starts with exposing high-textured dense missing samples at a fine scale. Then, by updating model parameters, this process continues and finally terminates by retrieving lowpass sparse samples at a coarse scale. The advised variable-scale windowing in comparison to fixed window speeds up the convergence rate of Algorithm~\ref{RecoveryAlgorithm} without sacrificing recovery quality. Also, this mechanism controls better error propagation during recursion. In Lemma 1, we prove the algorithm quickly converges after a few generations for an image with arbitrary degree of texture.

\newtheorem{Lemma1}{Lemma}
\begin{Lemma1}[Convergence guarantee]\label{Lem1}
Let patches of the image $\mathbf{I}=[i_{i,j}]_{h\times w}$ be square of the length $b=8$. If $N_g$ denotes the number of CA generations, then, Algorithm~\ref{RecoveryAlgorithm} for recovering the image $\mathbf{I}$ guarantees to converge at most at three iterations, i.e. $N_g\leqslant 3$.
\end{Lemma1}

\begin{IEEEproof}
To prove, we consider the worst case, i.e. the image under reconstruction is a fully DC image as the possible sparsest signal to determine the upper bound. In such a case, for all $\lceil\frac{h}{b}\rceil\times\lceil\frac{w}{b}\rceil$ patches, we have $\eta=0$ which results in $\mathbf{P}_u=\mathbf{P}_{\rm UVL}$, and the random and nonuniform patterns equal $\mathbf{P}_r=\mathbf{P}_n=\mathbf{O}$. Hence, $\mathbf{M}=\mathbf{P}_u$ for all patches. Let the live and dead cells be represented by white- and black- colored cells, respectively. For visualization, Fig.~\ref{Convergence} illustrates generations of the sampling mask matrix $\mathbf{M}_s$ for a given flat DC image of the dimension $32\times32$. It is important to note that, if $h$ and $w$ are not multiples of $b$, we can use zero padding. If the matrix $\mathbf{M}^{(\kappa)}_s$ denotes the sampling mask at generation $\kappa^{\rm th}$, then it is shown in Fig.~\ref{Convergence} that the number of dead cells at the $3^{\rm rd}$ generation is $N_d=\mathcal{Z}({\mathbf M}^{(3)}_s)=0$. This implies that $Ng=3$ for a completely DC image because the distance between two live cells is 7 pixels in both horizontal and vertical directions. For a given image with a certain degree of texture, the density of live cells in $\mathbf{M}_s$ is always more than or equal to the worst case, thus demonstrating $N_g\leqslant 3$.
\end{IEEEproof}

\newtheorem{Example1}{Example}
\begin{Example1}[Sampling and recovery]
\label{Ex1}
Figure~\ref{PipelineExample} shows the original images of Baboon, Boat, and Lena for which the sampling masks, subsampled images, the intermediate, and final results of recovery step are visualized. These images have different texture averages sorted in descending order, for which the sampler automatically extracted $35.23$, $25.16$ and $18.12$ percent pixels, and the recoverer estimated the images with $\rm PSNRs$ of $29.38$, $29.43$ and $32.19$ decibels, respectively.
\end{Example1}

\subsection{Practical Considerations}
\label{PracticalConsiderationsSection}
In order to implement the proposed sampling and recovery algorithms in practice, we can consider a mechanism similar to the structure of image acquisition using Rice single-pixel camera \cite{duarte2008single}. However, one of the main differences is to utilize a small array of sensors but not a single photo-diode. This allows us to measure adaptively the local image content, sample in 2-D space, and recover the scene by the suggested CA-based recovery approach. The solution is to focus the physical scene on a Digital Micro-mirror Device (DMD) via a primary lens. The DMD itself is partitioned into non-overlapping $b\times b$ patches such as $8\times 8$ blocks. In a sequential line-scan manner and at each time, only one of the analog image patches is reflected to another lens by configuring mirror positions. Other mirrors reflect the scene on a black absorbent surface. This secondary lens focuses the light of that single patch on an $8\times 8$ CCD or CMOS sensor. After amplifying the signal and passing through an Analog to Digital Converter (ADC), the digitized sub-image, ${\mathbf R}$, is obtained. Afterwards, the texture of the patch is measured. Then, based on the proposed sampler, the sampling mask, ${\mathbf M}$, and the sub-sampled image, ${\mathbf S}$, are determined. This process is repeated until the generation of the binary mask as well as the sub-sampled image pertaining to the last patch. The overall sampling mask, ${\mathbf M}_s$, and corresponding sampled image, ${\mathbf I}_s$, can be stored in a memory card. In the decoding phase, the proposed recoverer gets ${\mathbf M}_s$ and ${\mathbf I}_s$ and finally restores the image scene, $\widehat{\mathbf {I}}_p$.

\section{Experiments}
\label{ExperimentalRes}
This section provides extensive experiments to support our findings. Algorithms were implemented in MATLAB and run on an Intel Core i7 2.2GHz laptop with 8GB RAM.

\subsection{Experimental Settings}
In order to evaluate the performance of our sampling/recovery framework and comparing it with other related approaches, we utilized various databases having different statistical characteristics. Image sets include the standard databases of NCID \cite{liu2011neighboring}, CCID \cite{olmos2004biologically}, UCID \cite{schaefer2003ucid}, Microsoft Research Cambridge Object Recognition Image Database\footnote{https://www.microsoft.com/en-us/download/details.aspx?id=52644}, and a collection of well-known test images such as Baboon, Cameraman, Lena, etc. Beside, we gathered a set of IR test images as well as a mega-pixel range images database called RCID \cite{taimori2016novel}, to more precisely investigate the efficiency of our algorithms. Some of these images are shown in Fig.~\ref{RCID_and_IR_Images}.
In order to evaluate recovery performance, we employed standard objective criteria of PSNR, SSIM \cite{wang2004image}, and Normalized Recovery Error (NRE) defined as ${\rm NRE}\triangleq{\parallel{\widehat{\mathbf {I}}_p}- \mathbf I \parallel}_{\ell_2}/{{\parallel \mathbf I \parallel}_{\ell_2}}$ as well as subjective evaluation.
We compared our sampling and recovery algorithms to related approaches including the intelligent MAR sampler suggested in \cite{yang2016high}, the scattered data recovery technique of spline interpolation and the state-of-the-art methods of BCS \cite{ji2008bayesian}, uHMT-NCS AMP \cite{shahrasbi2017model}, IMAT \cite{marvasti2012sparse}, IMATI \cite{zayed2014new}, ISP \cite{sadeghi2016iterative}, and TSCW-GDP-HMT \cite {sadeghigol2016model}.

We adjusted the set of parameters so that, in the suggested sampler, the PSNR is maximum for a given average rate. For tuning parameters, we used images of UCID database as validation set. To generate uniform matrices in the uniform sampler, different regular and periodic lattices with various punched points may be used. We examined a set of such configurations and finally selected the patterns introduced in Fig.~\ref{UniformPatterns}, which optimize the aforementioned objective criterion. We also set parameters of CA-based recoverer and its post-processing stage to obtain a sub-optimal objective PSNR performance criterion in dB. The influence of post-processing stage with appropriate parameters of $\sigma_f$ and $\rho$ in the CA-based recovery algorithm is to improve PSNR up to about $\rm 0.5 dB$. The standard deviation of smoother $\sigma_f$, meanwhile, has negligible impact on performance. The number of punched points in the uniform sampler, the coefficients $c$ and $d$ in the random sampler, and the threshold $\tau$ in the nonuniform sampler control the sampling rate in the suggested algorithm. On one hand, for natural image patches, we have often $\eta\leq45$, thus considering short intervals for such textures. This also means the number of punched points is often less than or equal to 4. On the other hand, the main usage of the coefficient $d$ is to make the formula acceptable for dense images, i.e. high values of $k$. Therefore, the influential factors for increasing or decreasing the adaptive sampling rate are the coefficient $c$ and the threshold $\tau$.

\subsection{Performance on Standard Databases}
\begin{figure}[!t]
\centering
\includegraphics[width=5cm]{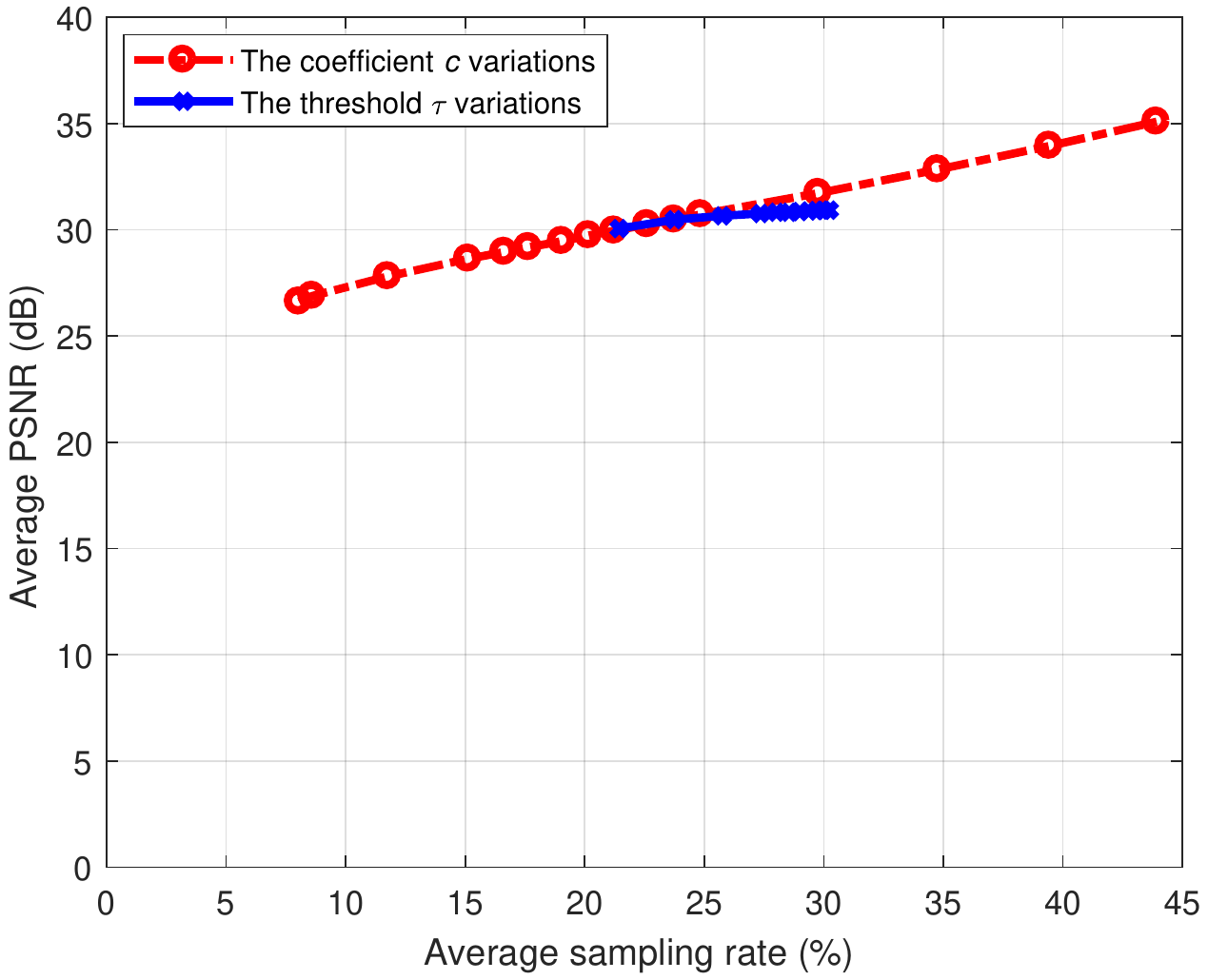}
\caption{Average PSNR ($\rm dB$) vs average sampling rate ($\%$) on images listed in Table~\ref{PerOnWellKnownImages}.}
\label{AveragePSNR_vs_AverageSamplingRate}
\end{figure}

\begin{figure}[!t]
\centering
\includegraphics[width=5cm]{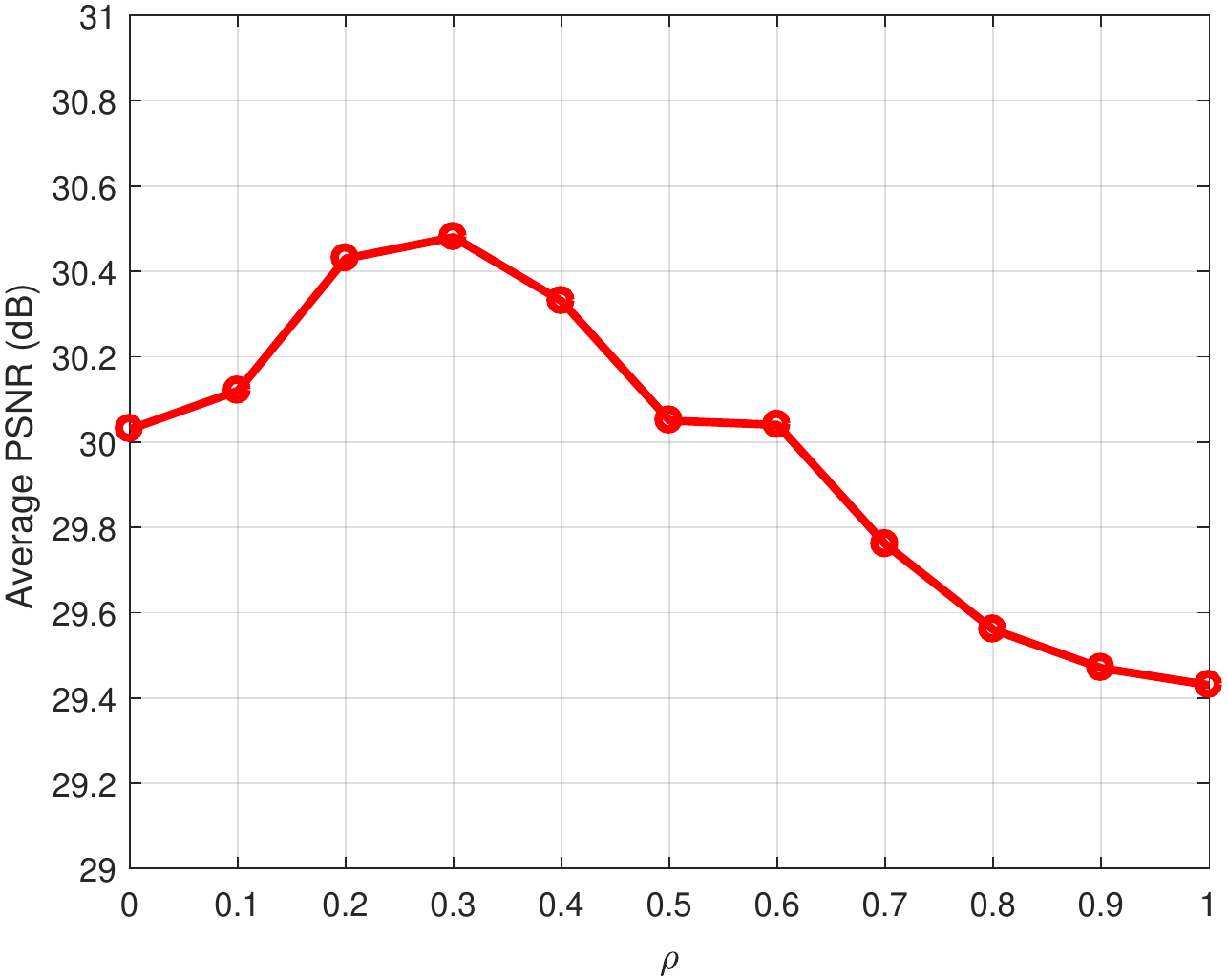}
\caption{The influence of the coefficient $\rho$ on recovery performance for images listed in Table~\ref{PerOnWellKnownImages}.}
\label{AveragePSNR_vs_Rho}
\end{figure}

Here, we evaluate the performance of our sampling/recovery method on 21 well-known test images of the size $512\times 512$ and four standard databases. The adaptive sampling rate measured by sampler and recovery quality on image processing test images are tabulated in Table~\ref{PerOnWellKnownImages}. For an average $23.75\%$ dynamic sampling rate, the mean of $\rm PSNR$ was $\rm 30.48 dB$. For variations $0\leq c\leq 3$ and $0\leq \tau \leq 1$, the curves of the average PSNR (dB) vs the average sampling rate (\%) on 21 well-known images are depicted in Fig.~\ref{AveragePSNR_vs_AverageSamplingRate}. The curves demonstrate that the sensitivity of our algorithm is low for variations $c$ and $\tau$. Figure~\ref{AveragePSNR_vs_Rho} also shows average PSNR (dB) vs $\rho$. In this figure, the coefficient $\rho=0$ means the recovery without applying post-processing step and the optimal value is $\rho^*=0.3$, thus demonstrating about $\rm 0.5 dB$ gain.

\begin{figure}[!t]
\centering
\includegraphics[width=\linewidth]{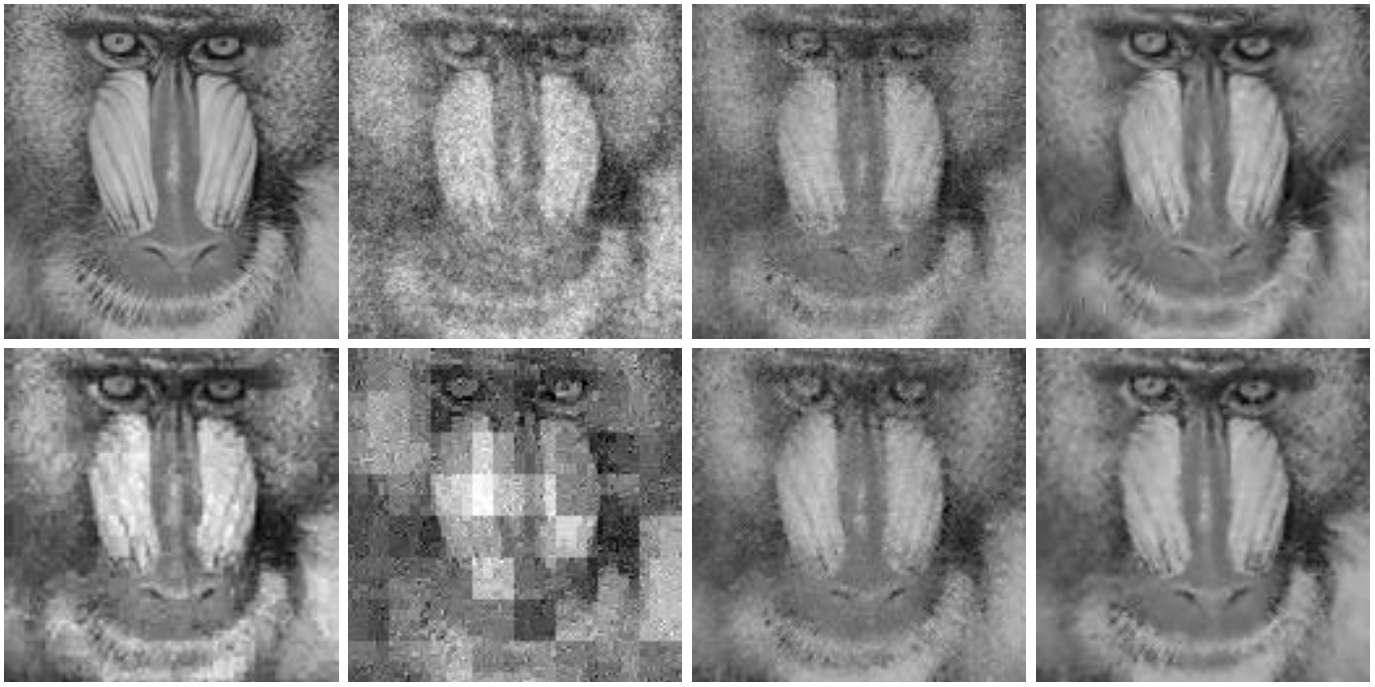}
\caption{PSNR for different methods in dB. From top to down and left to right, the original image, recovered images obtained by uHMT-NCS AMP (24.15), ISP (19.46), BCS (18.58), IMAT (24.25), IMATI (24.95), TSCW-GDP-HMT (26.72), and the proposed algorithm (27.33), respectively.}
\label{RecoveredBaboon}
\end{figure}

\begin{figure}[!t]
\centering
\includegraphics[width=5cm]{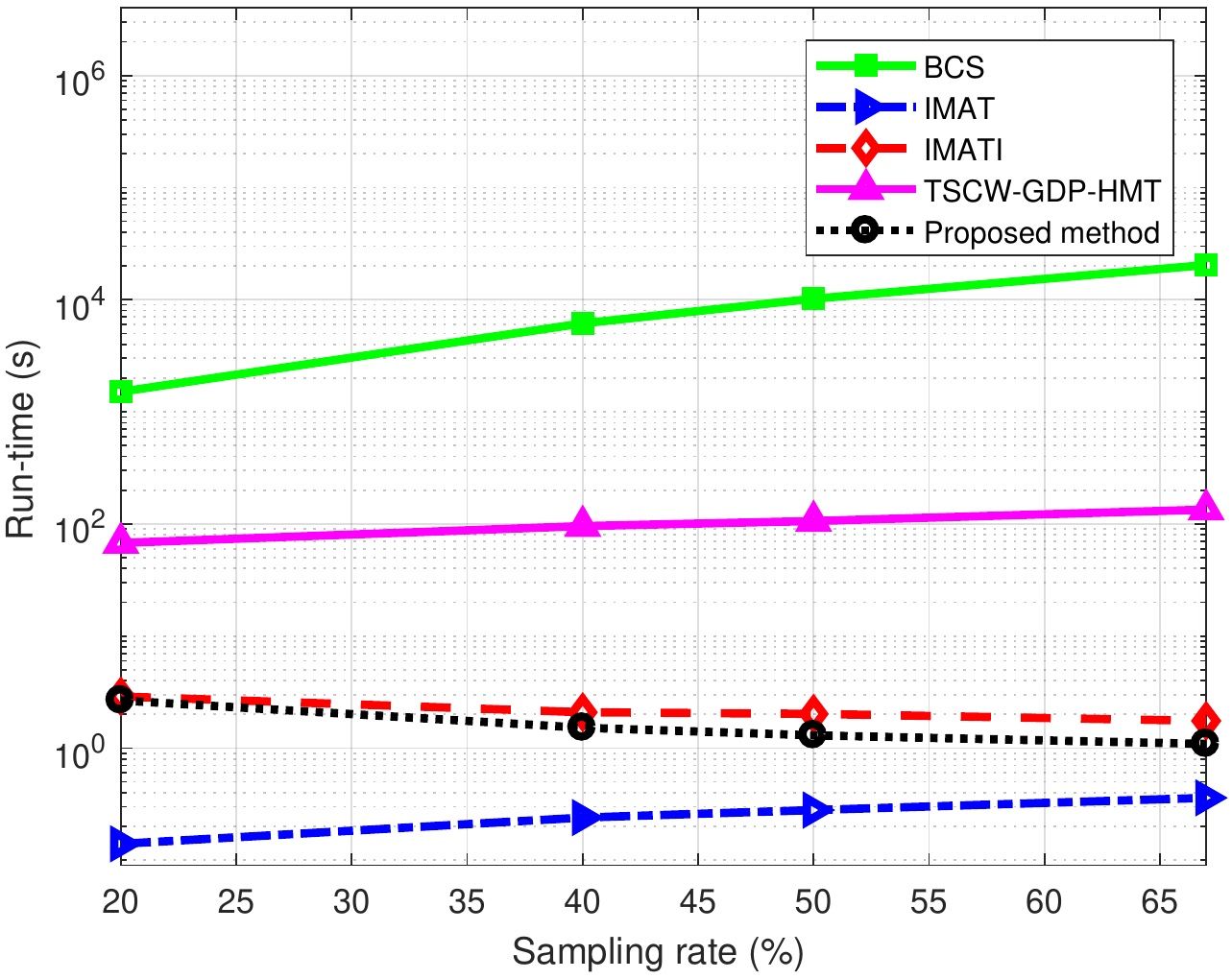}
\caption{The run-time (s) of different methods in terms of sampling rate ($\%$) on the image of Fig.~\ref{RecoveredBaboon}.}
\label{RunTime_Baboon}
\end{figure}

As an example, Fig.~\ref{RecoveredBaboon} illustrates recovery of Baboon image for compared state-of-the-art approaches. To be able to compare different algorithms of various complexities on the laptop, we reduced the image to the size $128\times128$. For all methods, the sampling rate was $36.62\%$, i.e. 6000 measurements. Based on PSNR, the proposed approach outperforms other methods. For BCS, IMAT, IMATI, TSCW-GDP-HMT, and the proposed algorithm, the average PSNR (dB) on all images with the size of $128\times128$ was 18.52, 23.34, 24.26, 28.54, and 27.99, respectively. Other methods such as ISP failed to compute the image restoration process. The average sampling rate determined by our algorithm was 34.3\%. For a fair comparison, we set this rate for competing algorithms, too. Although the average PSNR for TSCW-GDP-HMT algorithm is $\rm 0.55 dB$ more than the suggested method, its computational complexity is very high. To clarify this issue, Fig.~\ref{RunTime_Baboon} compares the run-time (s) of different methods in terms of sampling rate ($\%$) on the $128\times128$ Baboon image. As seen, the proposed method is slightly better than IMATI.



\begin{table}[!t]
\renewcommand{\arraystretch}{1.7}
\centering
\caption{Results of the proposed sampling/recovery method}
\centering
\label{PerOnWellKnownImages}
\centering
\resizebox{7.5cm}{!}{
\begin{tabular}{c|c||c|c c c}
\hline
\hline
\multirow{2}{*}{\textbf{Image number}} & \multirow{2}{*}{\textbf{Image name}} & \multirow{2}{*}{\textbf{Sampling rate (\%)}} & \multicolumn{3}{|c}{\textbf{Performance metrics}} \\ \cline{4-6}
& & & $\rm PSNR (dB)$ & $\rm SSIM$ & $\rm NRE$  \\
\hline
\hline
\multirow{1}{*}{1} & {Baboon} & 35.23	& 29.38	& 0.8814	& 0.0085	\\
 \hline
\multirow{1}{*}{2} & {Barbara} & 28.86 & 26.8	& 0.8431	& 0.0143 \\
\hline
\multirow{1}{*}{3} & {Boat} & 25.16	& 29.43	& 0.8227	& 0.0093 \\
\hline
\multirow{1}{*}{4} & {Butterfly} & 24.01	& 30.27	& 0.8356	& 0.0082 \\
\hline
\multirow{1}{*}{5} & {Cameraman} & 15.39	& 32.77	& 0.9295	& 0.0074 \\
\hline
\multirow{1}{*}{6} & {Einstein} & 18.09	& 31.26	& 0.7867	& 0.0082 \\
\hline
\multirow{1}{*}{7} & {Elaine} & 19.56	& 31.26	& 0.7613	& 0.0106 \\
\hline
\multirow{1}{*}{8} & {Fruits} & 16.6	& 31.13	& 0.8605	& 0.0056 \\
\hline
\multirow{1}{*}{9} & {Goldhill} & 25.51	& 30.4	& 0.8199 & 0.0107 \\
\hline
\multirow{1}{*}{10} & {Jetplane} & 18.59	& 31.36	& 0.91	& 0.0078 \\
\hline
\multirow{1}{*}{11} & {Lake} & 25.67	& 29.21	& 0.8324	& 0.0155 \\
\hline
\multirow{1}{*}{12} & {Lena} & 18.12	& 32.19	& 0.8779	& 0.0068 \\
\hline
\multirow{1}{*}{13} & {Livingroom} & 29.04	& 29.27	& 0.8288	& 0.0116 \\
\hline
\multirow{1}{*}{14} & {Owl} & 34.48	& 28.53	& 0.819	& 0.0115 \\
\hline
\multirow{1}{*}{15} & {Peppers} & 18.97	& 31.19	& 0.8464	& 0.0129 \\
\hline
\multirow{1}{*}{16} & {Pirate} & 24.62	& 29.99	& 0.8488	& 0.0095 \\
\hline
\multirow{1}{*}{17} & {Shack} & 34.1	& 27.09	& 0.8045	& 0.0234 \\
\hline
\multirow{1}{*}{18} & {Walkbridge} & 38	& 27.38	& 0.8279	& 0.0119 \\
\hline
\multirow{1}{*}{19} & {Woman-blonde} & 21.73	& 29.73	& 0.8229	& 0.0115 \\
\hline
\multirow{1}{*}{20} & {Woman-darkhair} & 11.55	& 36.69	& 0.9265	& 0.0053 \\
\hline
\multirow{1}{*}{21} & {Zelda} & 15.41	& 34.72	& 0.8915	& 0.0076 \\
\hline
\hline
\multicolumn{2}{c||}{\textbf{Average}} & \textbf{23.75} & \textbf{30.48}	& \textbf{0.8465} & \textbf{0.0104} \\
\hline
\hline
\end{tabular}}
\end{table}

Table~\ref{PerOnStandardDBs} reports the performance of our approach on NCID, CCID, UCID, and RCID databases, which are sorted in descending order of textural information; i.e. RCID and NCID databases have minimum and maximum texture averages, respectively. The average sampling rate acquired by the proposed dynamic sampler confirms this subject. Images of each image-set have also intra-database variability of textural information. However, we tuned parameters of our sampler so that dynamic sampling rate for each image doesn't approximately exceed 50\%, based on the limitation imposed in compressive sensing theory for real-world applications \cite{lustig2008compressed}. This leads to an acceptable average PSNR with a high variance, so that the range of PSNR (dB), [min(PSNR), max(PSNR)], is [17.39, 48.48], [21.91, 41.25], [18.41, 40.03], and [26.29, 44.35] for NCID, CCID, UCID, and RCID databases, respectively. From compression viewpoint, recovery performance in comparison to sensed samples is promising for all databases.

\begin{table*}[!t]
\renewcommand{\arraystretch}{1.7}
\centering
\caption{Performance of the proposed image sampling/recovery framework on four standard databases}
\centering
\label{PerOnStandardDBs}
\centering
\resizebox{12cm}{!}{
\begin{tabular}{c||c|c|c|c c c}
\hline
\hline
\multirow{2}{*}{\textbf{Database}} & \multirow{2}{*}{\textbf{Total images}} & \multirow{2}{*}{\textbf{Image dimensions}} & \multirow{2}{*}{$(\mu \pm \sigma^2)$ \textbf{of sampling rate (\%)}} & \multicolumn{3}{|c}{$(\mu \pm \sigma^2)$ \textbf{of performance metrics}} \\ \cline{5-7}
& & & & $\rm PSNR (dB)$ & $\rm SSIM$ & $\rm NRE$  \\
\hline
\hline
\multirow{1}{*}{NCID \cite{liu2011neighboring}} & 5150 & $256\times 256$ & $37.29\pm3.4$	& $26.57\pm26.04$	& $0.8355\pm0.0049$	& $0.02\pm9.56$\rm e-$5$ \\
\hline
\multirow{1}{*}{CCID \cite{olmos2004biologically}} & 1096 & $576\times 768$ & $26.68\pm1.35$ & $29.95\pm11.22$	& $0.849\pm0.0025$	& $0.0091\pm2.2$\rm e-$5$\\
\hline
\multirow{1}{*}{UCID \cite{schaefer2003ucid}} & 1338 & $384\times 512$ & $26.56\pm0.91$	& $28.58\pm9.1$	& $0.8654\pm0.0024$	& $0.0154\pm0.63$\rm e-$5$	 \\
\hline
\multirow{1}{*}{RCID \cite{taimori2016novel}} & 208 & $3456\times 5184$ & $12.24\pm0.54$	& $36.49\pm10.06$	& $0.9282\pm9.68$\rm e-4	& $0.0021\pm1.3$\rm e-6$$	 \\
\hline
\hline
\end{tabular}}
\end{table*}

\subsection{Test on Infra-Red Imaging Systems}
Infra-red imaging has a wide range of applications from surveillance and Intelligent Transportation Systems (ITSs) to medical imaging. These images are naturally sparse and of interest in the compressive sensing community to fabricate their low-cost CS-based imagers \cite{chen2013infrared}. In this experiment, we evaluate the performance of adaptive sampling/recovery framework on a test-set including 20 images grabbed by various near IR cameras for ITSs applications such as License Plate Recognition (LPR). Four representative IR images are shown in the first row of Fig.~\ref{RCID_and_IR_Images}. Specifically, for recovering images 1 to 4 in this figure, PSNRs(sampling rates) in terms of dB($\%$) were 46.68(2.08), 35.89(11.58), 42.04(4.41) and 38.86(6.57), respectively. Generally, all IR images were recovered by an average PSNR of $39.28$dB for the average adaptive sampling rate $7.16 \%$, thus demonstrating an excellent IR recovery quality of our scheme for the high sample reduction.

For BCS, IMAT, IMATI, TSCW-GDP-HMT, and the proposed algorithm, the average PSNR (dB) on all images with the size of $128\times 128$ was 28.44, 22.11, 26.34, 35.23, and 31.88, respectively. The average sampling rate determined by our algorithm was 15.37\%. For a fair comparison, we set this rate for competing algorithms. The performance of the proposed method is the second best.

\subsection{Influence of Uniform and Nonuniform Samples}
In this experiment, we considered two scenarios to evaluate the role of the proposed uniform and nonuniform samplers. In the first scenario, we omitted our proposed uniform sampler from the scheme of Fig.~\ref{SamplingScheme} and compared the resulting random/nonuniform sampler with the whole model. This can be done by setting $\mathbf{M}_u = \mathbf{O}$. In this case, for a fair comparison by preserving samples at an equal rate, we can manually increase the coefficient $c$ in the random sampler or decrease $\tau$ in the nonuniform sampler. Here, we only increased $c$ from the resulting random/nonuniform sampler. In this case, the average PSNR on 21 well-known images listed in Table~\ref{PerOnWellKnownImages} was $29.78$dB. In comparison to the result reported in Table~\ref{PerOnWellKnownImages}, a decrease of nearly $0.7$dB in PSNR is seen which demonstrates the importance of existing uniform sampler. It is noticeable that performance reduction in low-pass and periodic images like Cameraman and Barbara is much more than cluttered ones.

\begin{figure}[!t]
\centering
\includegraphics[width=7cm]{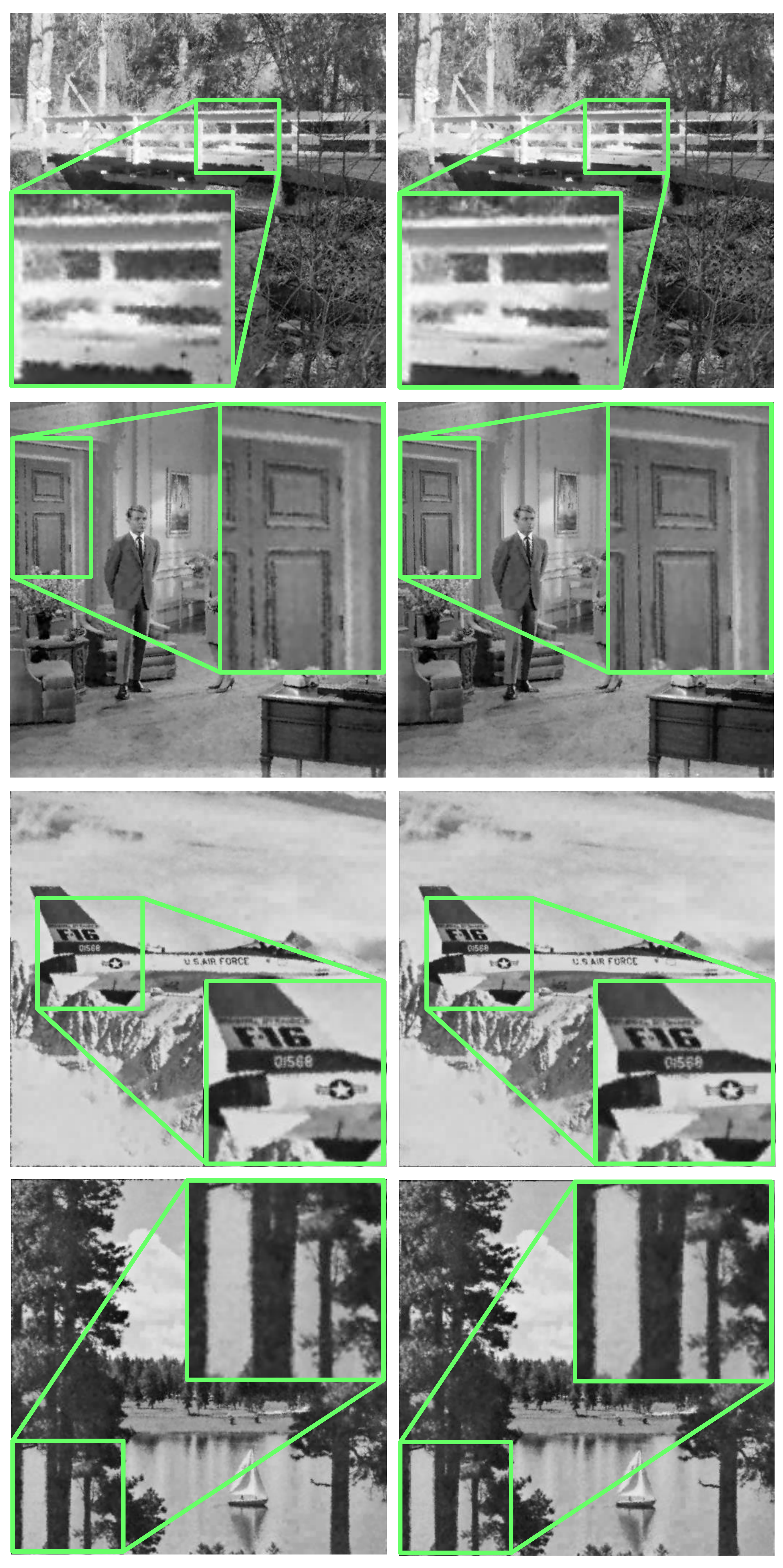}
\caption{From left to right and top to down, the reconstruction of Walkbridge, Livingroom, Jetplane, and Lake images without and with nonuniform sampler, respectively.}
\label{EffectOfNonuniformSampler}
\end{figure}

In the second scenario, the nonuniform sampler was neglected by setting $\tau = 1$. Similarly, we increased $c$ to equalize the rate of resulting uniform/random sampler with the complete scheme. The average PSNR was $30.52$dB, that is $0.04$dB more than the average of Table~\ref{PerOnWellKnownImages}. Although such an objective evaluation shows a little performance improvement in the absence of nonuniform sampler, different subjective evaluations narrate a better reconstruction of image structure such as edges in the presence of nonuniform sampler. For instance, Fig.~\ref{EffectOfNonuniformSampler} visualizes this phenomenon for four images in which reconstructed edges are sharper in the entire solution than the second scenario. Therefore, for approximately equal sampling rate and objective PSNR metric, the combined strategy has subjective superiority in performance.

\subsection{Recovery Performance under Fixed Sampling Structures}
This section compares the performance of different recovery approaches under fixed sampling structures. To this intent, we considered two sampling scenarios including conventional pure random and our measurement-adaptive samplers. For recovering scattered samples obtained from the proposed intelligent sampling stage as well as random sampling structure, various approaches such as scattered data interpolators can be employed. Therefore, in addition to modern sparse recovery techniques \cite{ji2008bayesian, marvasti2012sparse, zayed2014new, sadeghi2016iterative, sadeghigol2016model}, we applied the prominent scattered spline interpolator of spline to separately investigate the efficiency of our CA-based recoverer under the considered scenarios.

\begin{figure}[!t]
\centering
\includegraphics[width=7cm]{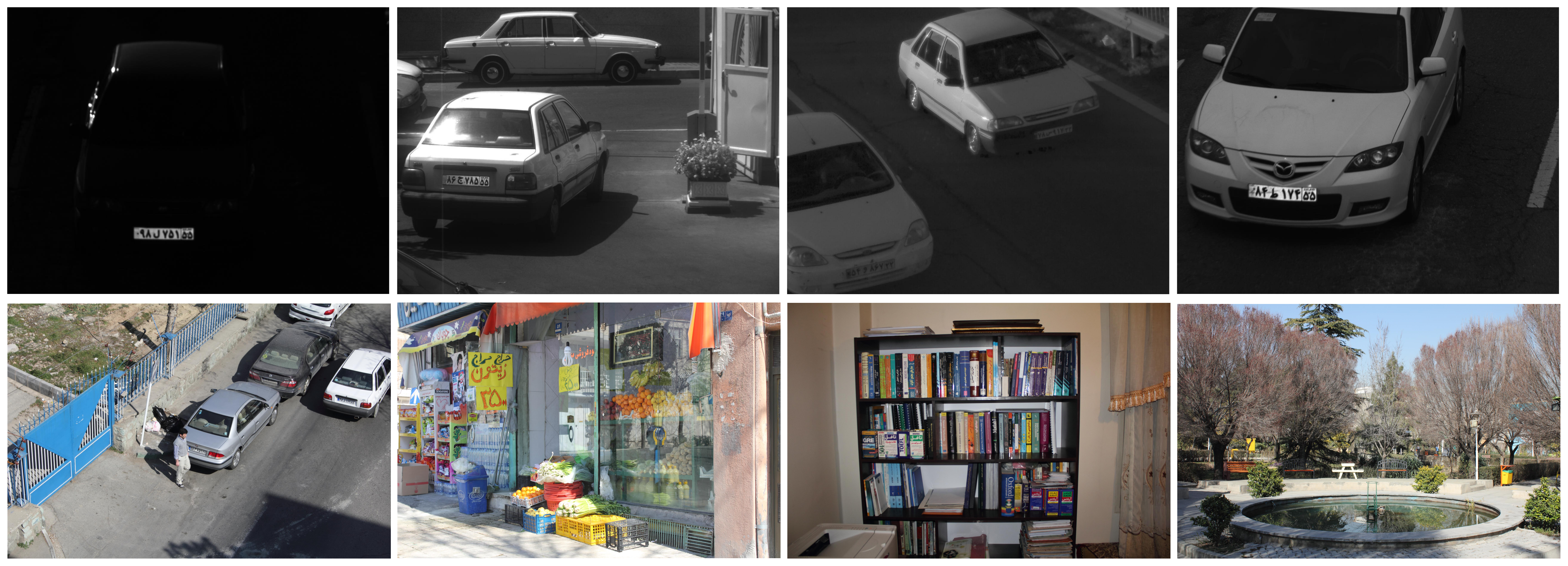}
\caption{From left to right, images in the first and the second rows indexed by numbers 1 to 4, show representative infra-red and RCID mega-pixel images \cite{taimori2016novel}, respectively.}
\label{RCID_and_IR_Images}
\end{figure}

In this experiment, we address the problem of image reconstruction of different scales. To do this, we selected 4 mega-pixel images from RCID database with the original size ${\rm min}(h, w)\times{\rm max}(h, w)=3456\times5184$ or vice versa. These images are shown in the second row of Fig~\ref{RCID_and_IR_Images}. To evaluate the efficiency of the mentioned approaches, by using the bicubic interpolation, we generated 6 down-sampled versions for each of which by a factor about $\downarrow{2}$ to finally construct a 7-level pyramid. Table~\ref{PerformanceUnderFixedSampling} compares the proposed CA-based recovery algorithm to modern reconstruction methods under random and adaptive sampling scenarios for original and scaled images. From lowest to highest resolution, i.e. in the direction of increasing correlation, the average dynamic sampling rates (\%) were 42.95, 38.63, 34.4, 30.21, 26.69, 22.37, and 15.59, respectively. For a fair comparison, we set the random sampling rate equal to that of the average dynamic rate determined by measurement-adaptive sampler. The bolded values show the best performance. Except for very low resolution levels, ISP, TSCW-GDP-HMT, BCS, and spline methods failed to yield viable results due to computational complexity of matrix operations or huge memory requirement. Because of heavy blockiness effects, BCS algorithm fails to reconstruct images in both the random measurement and the adaptive scheme especially in lowpass images. Generally, state-of-the-art results were obtained for configurations of measurement-adaptive sampler+spline for low-resolution and measurement-adaptive sampler+CA for high-resolution levels. This demonstrates the power of measurement-adaptive sampling scheme.

It is noticeable that IMAT and IMATI recovery methods reconstruct the whole image rather than block-wise processing. The sub-optimal parameters for adaptive thresholding technique of IMAT were $\alpha=0.5$, $\beta=50$, and $N_{\rm iteration}=30$. In ISP algorithm, DCT was utilized as sparsifying transform and hard thresholding function. We also set its sub-optimal parameters as $\gamma=0.4$, $\sigma_{\rm noise}=0$, $\tau_f=5$\rm e-6, $c=0.9$, $I=3$, and $\rm maxiter=300$. TSCW-GDP-HMT method uses Generalized Double Pareto (GDP) distribution for modeling signal sparsity. An inherent limitation of this algorithm is that the input image should be square with a length of multiples of 8. To tackle this problem and compare different approaches as fair as possible, we first considered the nearest length more than or equal to dimensions reported in Table~\ref{PerformanceUnderFixedSampling}, and then rescaled the recovered image to the real size. BCS algorithm utilizes Daubechies1 2-D wavelet at four decomposition levels. We also used MATLAB built-in function to implement spline interpolation.

\begin{table*}[!t]
\renewcommand{\arraystretch}{1.3}
\centering
\caption{Analyzing the performance of different recovery techniques in the case of two distinct sampling scenarios on images shown in the second row of Fig.~\ref{RCID_and_IR_Images}. For each scenario, the values in bold type show the best performance.}
\centering
\label{PerformanceUnderFixedSampling}
\centering
\resizebox{15cm}{!}{
\begin{tabular}{c||c|c c c c c c c|c c c c c c c}
\hline
\hline
\multirow{3}{*}{\textbf{Image number}} & \multirow{3}{*}{\textbf{Size at pyramid level}} & \multicolumn{14}{c}{${\rm PSNR (dB)}$} \\ \cline{3-16}
&  & \multicolumn{7}{c|}{\textbf{Random sampling scenario}} & \multicolumn{7}{c}{\textbf{Proposed measurement-adaptive sampling scenario}} \\ \cline{3-16}
&  & IMAT & IMATI & ISP & TSCW-GDP-HMT & BCS & Spline & CA & IMAT & IMATI & ISP & TSCW-GDP-HMT & BCS & Spline & CA \\
\hline
\hline
\multirow{7}{*}{1} & $56\times 80$ & 22.79 & 20.27  & 20.11 & \textbf{27.7} & 7.93 & 25.71 & 25.64 & 21.2 & 20.27 & 18.56 & 25.54 & 8.43 & \textbf{27.49} & 26.56\\
& $104\times 160$ & 22.88 & 24.1 & F & \textbf{29.56} & 8.77 & 26.44 & 26.02 & 21.03 & 21.73 & F & \textbf{28.55} & 8.94 & 28.37 & 27.15\\
& $216\times 320$ & 22.61 & 24.02 & F & F & F & F & \textbf{25.07} & 20.27 & 21 & F & F & F & F & \textbf{26.88}\\
& $432\times 648$ & 22.82 & \textbf{25.46} & F & F & F & F & 25.36 & 20.15 & 22.36 & F & F & F & F & \textbf{27.93}\\
& $864\times 1296$ & 23.84 & \textbf{27.82} & F & F & F & F & 27.81 & 19.87 & 23.26 & F & F & F & F & \textbf{30.27}\\
& $1728\times 2592$ & 24.74 & \textbf{31.57} & F & F & F & F & 31.49 & 18.59 & 24.19 & F &F & F & F & \textbf{33.3}\\
& $3456\times 5184$ & 22.45 & 35.31 & F & F & F & F & \textbf{35.33} & 15.58 & 26.48 & F & F & F & F & \textbf{36.7}\\
\hline
\multirow{7}{*}{2} & $56\times 80$ & 21.92 & 19.7 & 18.26 & \textbf{25.96} & 8.75 & 25.62 & 24.63 & 23.86 & 19.4 & 20.5 & 24.03 &  7.01 & \textbf{27.18} & 25.78\\
& $104\times 160$ & 22.82 & 23.72 & F & \textbf{27.3} & 7.63 & 26.06 & 25.27 & 23.9 & 23.87 & F & 26.23 & 8.79 & \textbf{27.84} & 26.63\\
& $216\times 320$ & 23.22 & 24.36 & F & F & F & F & \textbf{25.7} & 23.22 & 24.43 & F & F & F & F & \textbf{27.3}\\
& $432\times 648$ & 23.64 & 26.22 & F & F & F & F & \textbf{26.7} & 22.44 & 25.05 & F & F & F & F & \textbf{28.17}\\
& $864\times 1296$ & 24.22 & 27.74 & F & F & F & F & \textbf{28.1} & 21.49 & 25.39 & F & F & F & F & \textbf{29.78}\\
& $1728\times 2592$ & 24.58 & 29.61 & F & F & F & F & \textbf{30.27} & 19.19 & 25.25 & F & F & F & F & \textbf{32.19}\\
& $3456\times 5184$ & 21.92 & 31.51 & F & F & F & F & \textbf{32.77} & 14.73 & 25.44 & F & F & F & F & \textbf{35.16}\\
\hline
\multirow{7}{*}{3} & $56\times 80$ & 21.99 & 20.77 & 16.42 & \textbf{25.82} & 11.19 & 24.68 & 23.51 & 22.84 & 21.13 & 17.23 & 24.25 & 10.24 & \textbf{26.68} & 24.85\\
& $104\times 160$ & 23.21 & 22.8 & F & \textbf{28.7} & 10.35 & 25.55 & 24.33 & 22.33 & 23.64 & F & 27.47 & 9.24 & \textbf{27.55} & 26.51\\
& $216\times 320$ & 23.99 & \textbf{26.07} & F & F & F & F & 25.82 & 19.87 & 19.51 & F & F & F & F & \textbf{27.87}\\
& $432\times 648$ & 25.07 & \textbf{28.43} & F & F & F & F & 27.48 & 17.29 & 19.16 & F & F & F & F & \textbf{29.43}\\
& $864\times 1296$ & 26.47 & \textbf{31.77} & F & F & F & F & 30.43 & 16.04 & 20 & F & F & F & F & \textbf{32.3}\\
& $1728\times 2592$ & 27.36 & \textbf{35.47} & F & F & F & F & 34.41 & 14.91 & 21.73 & F & F & F & F & \textbf{35.71}\\
& $3456\times 5184$ & 24.37 & 35.9 & F & F & F & F & \textbf{36.76} & 13.94 & 26.33 & F & F & F & F & \textbf{37.06}\\
\hline
\multirow{7}{*}{4} & $56\times 80$ & 23.11 & 19.12 & 20.01 & \textbf{27.84} & 7.88 & 26.91 & 24.95 & 22.11 & 19.64 & 20.84 & 25.81 & 9.71 & \textbf{27.58} & 26.08\\
& $104\times 160$ & 22.97 & 23.64 & F & \textbf{28.35} & 8.26 & 26.29 & 25.4 & 21.14 & 22.61 & F & 27.55 & 8.34 & \textbf{27.68} & 26.55\\
& $216\times 320$ & 22.72 & 24.49 & F & F & F & F & \textbf{25.36} & 20.67 & 22.85 & F & F & F & F & \textbf{26.92}\\
& $432\times 648$ & 22.35 & 24.54 & F & F & F & F & \textbf{25.25} & 20.27 & 23.04 & F & F & F & F & \textbf{27.34}\\
& $864\times 1296$ & 21.97 & 24.03 & F & F & F & F & \textbf{25.07} & 19.64 & 23.14 & F & F & F & F & \textbf{27.59}\\
& $1728\times 2592$ & 21.74 & 24.33 & F & F & F & F & \textbf{25.88} & 18.57 & 23.23 & F & F & F & F & \textbf{28.99}\\
& $3456\times 5184$ & 20.14 & 25.15 & F & F & F & F & \textbf{27.3} & 16.32 & 23.39 & F & F & F & F & \textbf{31.2}\\
\hline
\hline
\end{tabular}}
\end{table*}

\subsection{Evaluation under Fixed Recovery Algorithms}
\begin{figure}[!t]
\centering
\includegraphics[width=5cm]{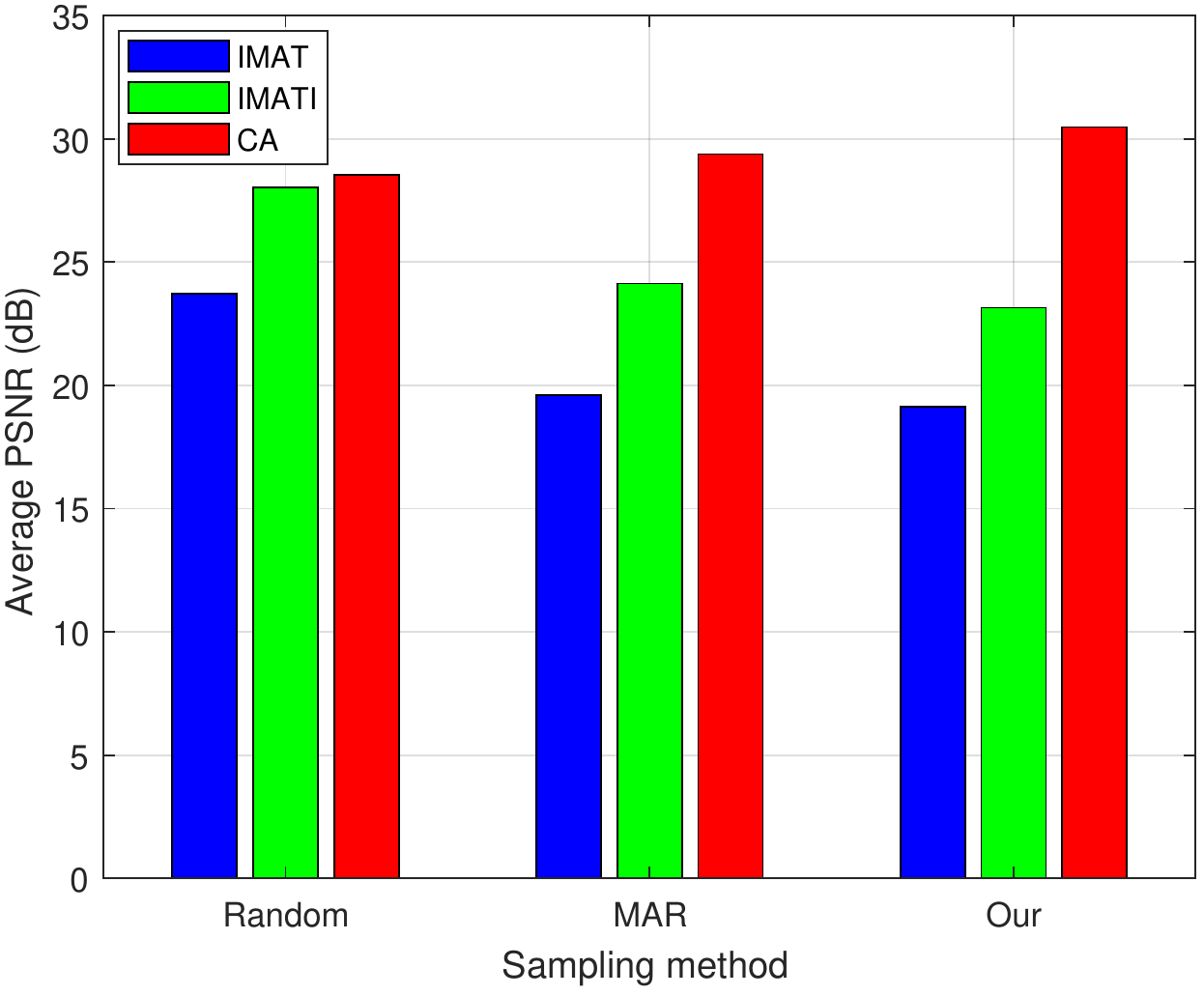}
\caption{The comparison of recovery performance for different sampling approaches.}
\label{DifferentSamplersUnderFixedRecoveryAlgorithms}
\end{figure}

In this experiment, we evaluate the performance of three sampling methods including the pure random, the state-of-the-art dynamic MAR sampling approach~\cite{yang2016high}, and the suggested sampler under fixed recovery algorithms. We utilized 21 well-known test images of the original size $512\times512$ for performance evaluation. As known in Table~\ref{PerformanceUnderFixedSampling}, only IMAT, IMATI and the proposed CA-based recoveres can restore the images of such dimensions. Hence, these recovery algorithms were considered to be able to recover subsampled images. MAR sensing matrix is basically generated from the low-resolution versions of images with the size $128\times128$. For a fair comparison with this method, we created subsampled images from Hadamard product of MAR masks and original images in recovery side. Figure~\ref{DifferentSamplersUnderFixedRecoveryAlgorithms} depicts average PSNR criterion in dB for different sampling/recovery configurations. The sampling rate for all samplers is the same and equal to 23.75\%. The first and the second ranks are belonging to the proposed measurement-adaptive sampler+CA recoverer and MAR sampler+CA recoverer, respectively. It is noticeable that the performance of IMAT and IMATI recovery algorithms under the pure random sampling strategy are better than other state-of-the-art samplers due to their nature.

\subsection{Recovery Robustness in the Presence of Noise}
\begin{figure}[!t]
\centering
\includegraphics[width=5cm]{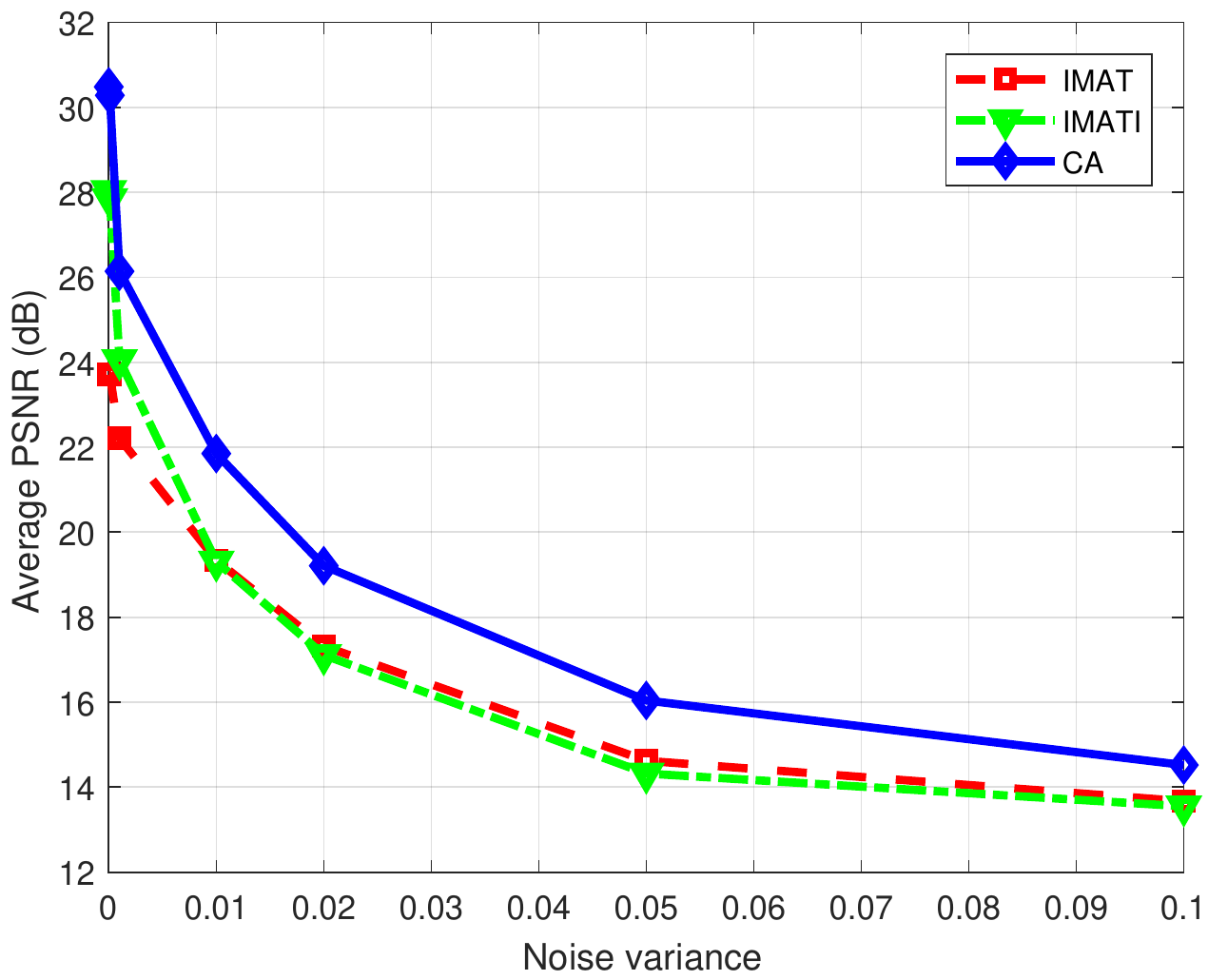}
\caption{Average PSNR (dB) vs the noise variance, $\sigma_{\rm AWGN}^2$, for the case of noisy measurements.}
\label{PlotOfPSNR_vs_NoiseVariance}
\end{figure}

In order to evaluate the robustness of the proposed CA-based recovery algorithm in the presence of noise, we can model the noisy subsampled image, $\overline{\mathbf{I}}_s$, as
\begin{equation}
\label{SampledPatch}
\overline{\mathbf{I}}_s = \mathbf{I}_s + (\mathbf{M}_s \odot \mathbf{N}),
\end{equation}
where matrices $\mathbf{I}_s$, $\mathbf{M}_s$, and $\mathbf{N}$ denote the noise-free subsampled image, the binary sampling mask, and Additive White Gaussian Noise (AWGN) with the distribution of $\mathcal{N}(0,\sigma_{\rm AWGN}^2 )$, respectively. We then applied the normalized version of $\overline{\mathbf{I}}_s$ between 0 and 1 to CA recoverer. Figure~\ref{PlotOfPSNR_vs_NoiseVariance} plots average PSNR in dB vs the noise variance, $\sigma_{\rm AWGN}^2$, for 21 well-known test images of the size $512\times 512$. In this figure, $\sigma_{\rm AWGN}^2=0$ means the noise-free case. The suggested method shows better performance in terms of the noise variance than IMAT and IMATI approaches. All curves exponentially decay by increasing $\sigma_{\rm AWGN}^2$. As mentioned before, other algorithms fail to recover these images due to data dimensionality.

\subsection{Performance of Joint Sampling/Recovery Frameworks}
\begin{figure}[!t]
\centering
\includegraphics[width=7cm]{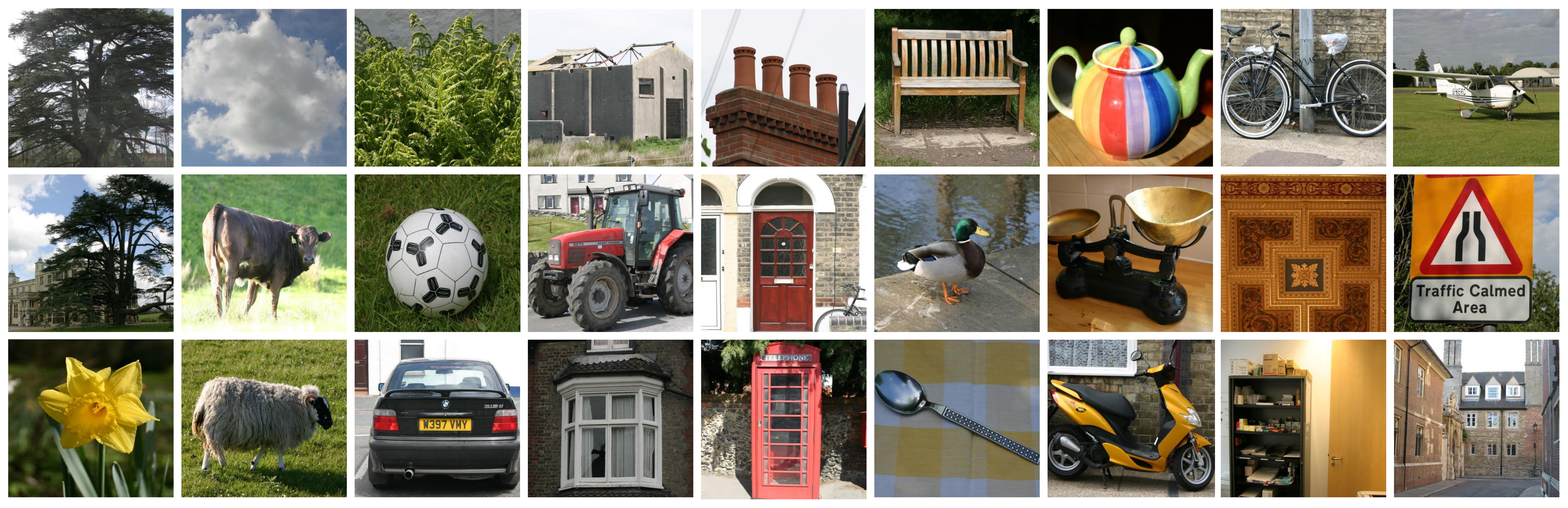}
\caption{From top to down and left to right, test samples of Microsoft Database indexed by numbers 1 to 27.}
\label{MicrosoftDB}
\end{figure}

\begin{figure}[!t]
\centering
\includegraphics[width=6cm]{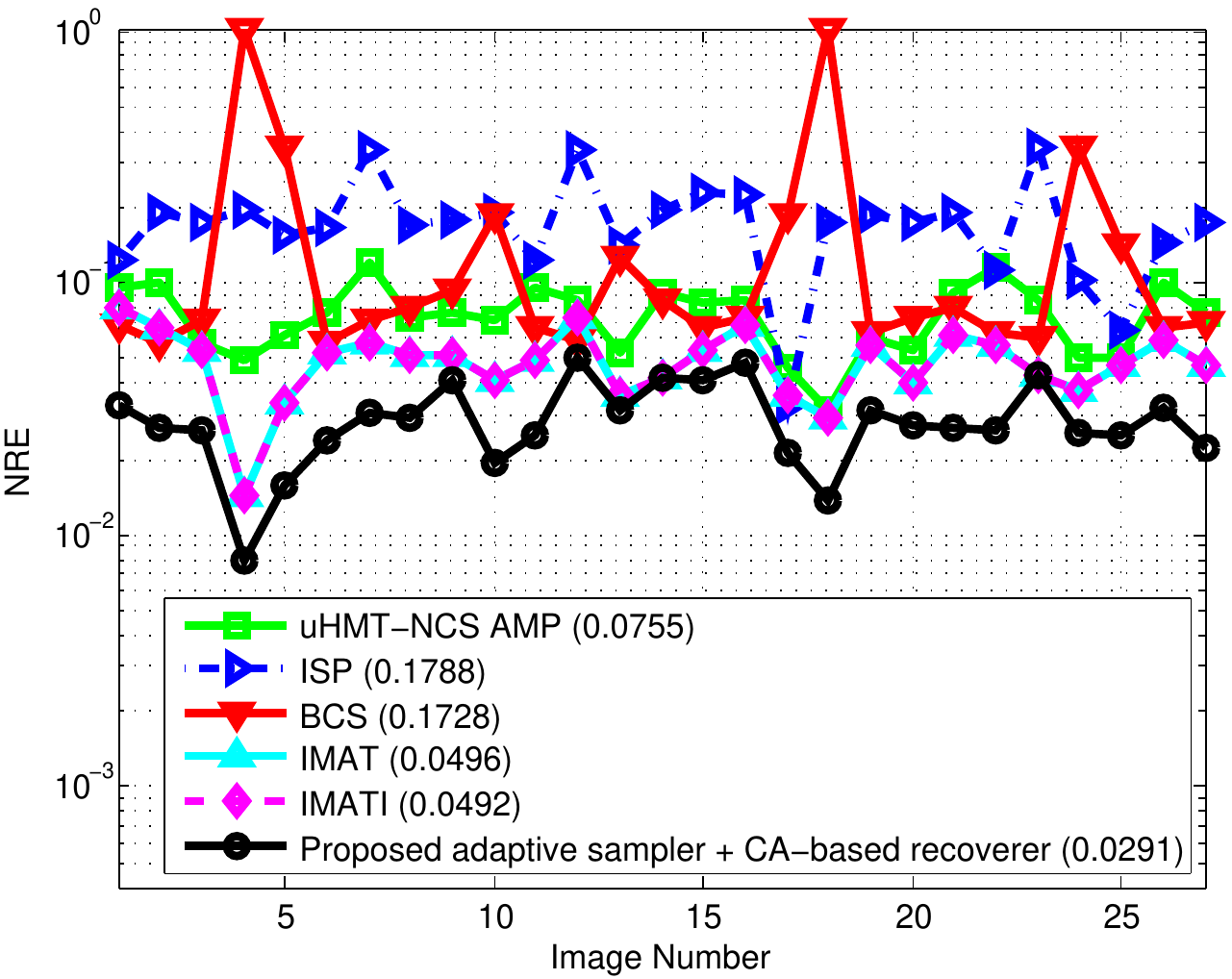}
\caption{Normalized Recovery Error (NRE) of different approaches for the images labeled in Fig.~\ref{MicrosoftDB}.}
\label{NRE_Microsoft}
\end{figure}

In practice, design of sampling structure and recovery stage are closely related to each other. Here, we jointly compare the performance of the proposed sampling/recovery with other state-of-the-art frameworks \cite{shahrasbi2017model, marvasti2012sparse, zayed2014new, sadeghi2016iterative, ji2008bayesian}. To do this, we collected 27 randomly selected samples from Microsoft Object Recognition Database as shown in Fig.~\ref{MicrosoftDB}, which were already employed in \cite{shahrasbi2017model}. At first, the randomly chosen images were cropped for highlighting salient objects in the scene, and then their gray-scale versions were resized to $128\times128$ dimensions using the bicubic interpolation to be able to compare different algorithms of various complexity. Figure~\ref{NRE_Microsoft} plots NRE vs image number as indexed by numbers 1 to 27 in Fig.~\ref{MicrosoftDB} for competing methods. The average $\rm NREs$ are given in legend parentheses. As shown, the average NRE of the proposed scheme is lower than other methods. BCS failed to recover the smooth images 4 and 18, whereas our scheme retrieved them with minimum recovery error. For a fair comparison, we set the sampling rate of competing methods the same as the average rate $32.44\%$ obtained adaptively by our sampler.

\section{Conclusion}
\label{ConclusionSection}
This paper suggested a measurement-adaptive sparse image sampling and recovery framework. The proposed sampler adaptively measures the required sampling rate and accordingly determines sample positions by a combined uniform, random, and nonuniform sampling mechanism. Our idea originated from the fact that natural images may have lowpass, bandpass, highpass, or sparse components depending on the scene under view. Therefore, a measurement-adaptive mechanism was advised to trap informative samples. Unlike Gaussian measurement sensing scenario, the proposed sparse coding style does not need all samples in advance.

In the recovery phase, we modeled the problem by a cellular automaton machine. The suggested algorithm converges always at a few generations. Low computational burden and memory usage are two main advantages of this algorithm in comparison to sophisticated techniques of ISP \cite{sadeghi2016iterative}, TSCW-GDP-HMT \cite{sadeghigol2016model}, and BCS \cite{ji2008bayesian}, especially in reconstruction of mega-pixel range imaging such as remote sensing. In CA-based recovery algorithm, updating rule for reviving dead cells is done based on a simple weighted averaging. As a future work, more precise predictors can be utilized for this purpose. Also, the suggested sampling/recovery pipeline can be generalized to other sparsifying transform domains like wavelets. Extensive tests on standard image data-sets, infra-red, and mega-pixel imaging show the capabilities of proposed technique for practical implementations of compressively sampled imaging systems.

\appendix  
Based on JPEG standard \cite{luo2010jpeg}, the quantization matrix corresponding to the texture $\eta$ can be formulated as
\begin{equation}
\label{QuantizationMatrix}
\mathbf{Q}_{\eta}= \max_{1\le i,j\le b}(\lfloor \frac{s_\eta}{100} {{\mathbf Q}}_{r} + \frac{1}{2}\mathbf J \rfloor, \mathbf J),
\end{equation}
in which the matrix $\mathbf J$ denotes a ${b\times b}$ all-one matrix. Also, the scaling factor, $s_{\eta}$, and the reference quantization table, ${\mathbf{Q}_{r}}$, are determined by
\begin{equation}
\label{ScalingFactor}
{s_{\eta}}=\left\{ \begin{array}{cc}
\frac{5000}{\eta}, & 0^+\leq\eta<50  \\
2(100-\eta), & 50\leq\eta\leq100 \end{array}
\right.,
\end{equation}
and

${\mathbf{Q}_{r}}=\left( \begin{array}{cccccccc}
{\rm 16} & {\rm 11} & {\rm 10} & {\rm 16} & {\rm 24} & {\rm 40} & {\rm 51} & {\rm 61} \\
{\rm 12} & {\rm 12} & {\rm 14} & {\rm 19} & {\rm 26} & {\rm 58} & {\rm 60} & {\rm 55} \\
{\rm 14} & {\rm 13} & {\rm 16} & {\rm 24} & {\rm 40} & {\rm 57} & {\rm 69} & {\rm 56} \\
{\rm 14} & {\rm 17} & {\rm 22} & {\rm 29} & {\rm 51} & {\rm 87} & {\rm 80} & {\rm 62} \\
{\rm 18} & {\rm 22} & {\rm 37} & {\rm 56} & {\rm 68} & {\rm 109} & {\rm 103} & {\rm 77} \\
{\rm 24} & {\rm 35} & {\rm 55} & {\rm 64} & {\rm 81} & {\rm 104} & {\rm 113} & {\rm 92} \\
{\rm 49} & {\rm 64} & {\rm 78} & {\rm 87} & {\rm 103} & {\rm 121} & {\rm 120} & {\rm 101} \\
{\rm 72} & {\rm 92} & {\rm 95} & {\rm 98} & {\rm 112} & {\rm 100} & {\rm 103} & {\rm 99} \\ \end{array}
\right)$.


%

%

\section*{Acknowledgements}
This Postdoc research was jointly sponsored by Iran National Science Foundation (INSF) and ACRI of Sharif University of Technology under agreement numbers 95/SAD/47585 and 7000/6642, respectively. The authors would like to thank Dr Z. Sadeghigol who provided TSCW-GDP-HMT codes for comparison. We also thank Prof A. Amini, Mr A. Esmaeili, and other researchers in Signal Processing and Multimedia Lab for their priceless comments.

\ifCLASSOPTIONcaptionsoff
  \newpage
\fi



%
%
%

\bibliographystyle{IEEEtran}
%

\bibliography{bare_jrnl}

%

%






\end{document}